\documentclass[letterpaper]{article}
\usepackage[preprint]{aaai2027}
\usepackage[hyphens]{url}
\usepackage{graphicx}
\usepackage{natbib}
\usepackage{caption}
\usepackage{booktabs}
\usepackage{amsmath,amssymb,amsfonts,amsthm}
\usepackage{xcolor}
\usepackage{algorithm}
\usepackage{algpseudocode}

\newtheorem{theorem}{Theorem}
\newtheorem{corollary}{Corollary}
\newtheorem{proposition}{Proposition}

\newcommand{\ours}{F-CS-WAE}
\newcommand{\zc}{z_{\mathrm{c}}}
\newcommand{\zs}{z_{\mathrm{s}}}
\newcommand{\muc}{\mu_{\mathrm{c}}}
\newcommand{\mus}{\mu_{\mathrm{s}}}
\newcommand{\Sph}{\mathbb{S}}
\newcommand{\R}{\mathbb{R}}
\newcommand{\MMD}{\mathrm{MMD}}
\newcommand{\KL}{\mathrm{KL}}
\newcommand{\CE}{\mathrm{CE}}
\newcommand{\SCauchy}{\mathrm{SCauchy}}

\newcommand{\Dinter}{\Delta_{\mathrm{inter}}}

\title{Marginal Matching Does Not License Factorized Sampling:\\
Auditing Conditional Style Leakage in Factorized Generative Models}

\author{
  Duong Bach,\quad Hai Nguyen Hong,\quad Cuong Do%
  \thanks{Code: \protect\url{https://github.com/DngBack/CS-WAE}}
}
\affiliations{
  Smart Health Center (VISHC), VinUniversity, Ha Noi, Viet Nam
}

\begin{document}
\maketitle

\begin{abstract}
Factorized generative models regularize a style latent $\zs$ toward a fixed
prior with a marginal statistic, $q(\zs)\approx\mathcal N(0,I)$, and treat
the result as a certificate that $\zs$ carries no class information. The
certificate does not hold. Matching the marginal constrains nothing about
the class conditionals $q(\zs\mid y)$, so $\zs$ can be exactly Gaussian in
aggregate while remaining maximally informative about the label. We show
this gap is one of four terms in an exact decomposition of what a factorized
sampler must match, and that eliminating it is necessary but not sufficient
for the sampling procedure the factorization exists to support. Empirically,
a case-study model and four reference latent baselines with near-zero global
MMD all permit a linear probe to recover the label from $\zs$ at
$74$--$100\%$ (chance $10\%$); our case-study
model reaches $99.15\%$ clustering accuracy while externally evaluated
class-conditional generation succeeds $16\%$ of the time. The leakage
survives six one-at-a-time perturbations of capacity, curriculum, prior
geometry, and supervision on two datasets. Four remedies attacking the leakage through
different mechanisms span $21$--$46\%$ probe recovery while leaving the
within-class dependence proxy essentially unchanged. A post-hoc conditional
prior raises external generated-class accuracy to $0.97$ on MNIST without
retraining but reaches only $0.41$ on CIFAR-10; an empirical style bank
reaches $0.88$ on CIFAR-10. The methodological point: no divergence computed
on $q(\zs)$ alone can certify $\zs\perp y$, and reporting marginal statistics
does not verify the property practitioners claim from them.
\end{abstract}

\section{Introduction}
\label{sec:intro}

Matching an aggregate posterior to a Gaussian is a statement about a
marginal. Class-invariant style is a statement about a family of
conditionals. The two are different claims, and the first does not imply the
second: $q(\zs)$ can be exactly $\mathcal N(0,I)$ while every $q(\zs\mid
y{=}k)$ occupies its own region of latent space, so long as the regions
average out. A regularizer that only ever sees $q(\zs)$ cannot detect this,
and therefore cannot prevent it.

The gap has teeth because of how factorized models generate. During
training the decoder only ever receives $(\zc,\zs)$ drawn from the same
image. If $q(\zs\mid y{=}k)$ differs across classes, and nothing in the
objective stops it, the decoder is free to learn class-specific interactions
between the two codes. At generation time a class prior supplies $\zc$ for
class $k$ while $\zs$ comes from the marginal, which averages over all
classes. The resulting pair is one the decoder never saw. We call the
failure \emph{conditional style leakage}.

What makes it dangerous is that the metrics a practitioner would check are
all blind to it. Reconstruction is measured on the posterior manifold, where
$\zs$ comes from a real image and the learned interaction is exactly the one
being exercised, so it looks fine. Global MMD is a marginal statistic by
construction. Clustering accuracy reads only $\zc$. A model can therefore
pass every standard diagnostic and still fail completely at the one
capability the factorization was built to provide.

Our starting point is that this is not a defect of any particular loss but a
structural property of factorized sampling. We give an exact decomposition
of the mismatch such a sampler incurs (Theorem~\ref{thm:decomp}) into four
non-negative terms: global style-prior mismatch, conditional style leakage,
semantic-prior mismatch, and within-class dependence between the two codes.
Valid sampling requires all four to vanish, so removing the leakage is
necessary but not sufficient, and the latent mismatch upper-bounds the
image-space mismatch but not conversely. That asymmetry is what makes a
decoder-level intervention necessary rather than redundant.

We then audit a case-study model and four reference latent baselines with
finite-sample proxies for the four terms, and use \ours{} (Factorized
Class-Structured Spherical Cauchy WAE) as
a concentrated case study. We use it to exhibit and repair the failure, not
to argue it beats the baselines it is compared against. The four reference
families share one diagnostic architecture and budget, while \ours{} is
evaluated in its native configuration, so the cross-model table supports
within-model auditing rather than a ranking. \ours{} is label-guided by
design; the proposed per-class style MMD reuses the labels already required
by its semantic objective and therefore adds no annotation requirement.

\paragraph{Contributions.}
(1) An auditing framework for factorized generative models that separates
marginal prior fit, conditional leakage, semantic-prior mismatch, and
within-class dependence, supported by a structural KL decomposition. (2) A
conditional audit protocol built from finite-sample proxies for those terms,
applied to a case-study model and reference latent baselines. (3) A
decoder-level validation study showing when information detected by the audit
is actually used during recombination, together with a
six-way robustness check on two datasets, showing no single design choice
accounts for the leakage while its magnitude is strongly capacity-dependent
on CIFAR-10. (4) An evaluation of representation- and sampling-level
repairs, including a negative transfer result.

\section{Related Work}

We organize prior work by a single question: does the method certify
conditional invariance $\zs\perp y$, or only a statistic compatible with
$\zs\not\perp y$?

VAEs~\citep{kingma2013autoencoding} impose a per-example KL toward a fixed
prior; WAEs~\citep{tolstikhin2017wasserstein} relax this to a divergence
between the aggregated posterior and the prior, computed with kernel
two-sample tests~\citep{gretton2012kernel} or projected
distances~\citep{kolouri2018sliced}. The VAE penalty is not simply an
aggregate regularizer in disguise: averaged over the data it splits exactly
into $\mathbb E_{q(x)}[\KL(q(z\mid x)\|p(z))] = I_q(X;Z) +
\KL(q(z)\|p(z))$~\citep{hoffman2016elbo}, an information-capacity term plus
an aggregate-mismatch term of the kind WAE-MMD targets directly. Neither
certifies $I(\zs;y)=0$. The capacity term bounds what $Z$ encodes about $X$
in general, not about the label, and the aggregate term is precisely the
marginal statistic Proposition~\ref{prop:nocert} shows is insufficient.

$\beta$-VAE~\citep{higgins2017beta}, $\beta$-TCVAE~\citep{chen2018isolating},
FactorVAE~\citep{kim2018disentangling}, and DIP-VAE~\citep{kumar2018dipvae}
penalize total correlation or aggregate moments. These constrain dependence
among latent \emph{dimensions}, not between a designated style code and the
label. More generally, unsupervised disentanglement is not identifiable
without inductive biases in the model or data~\citep{locatello2019challenging};
weakly supervised pairs can supply such a bias~\citep{locatello2020weakly},
but a bias toward factorization is not itself a certificate of the particular
invariance $\zs\perp y$. Semi-supervised VAEs~\citep{kingma2014semi} and
CVAE~\citep{sohn2015learning} condition generation on the label, sidestepping
the need for $\zc$ to encode identity, but do not by themselves test whether
a residual latent remains label-informative.

Content--style separation predates deep generative models
\citep{tenenbaum2000separating}. Modern split-latent models use class labels,
grouped observations, pairwise similarity, mutual-information penalties, or
latent optimization to divide specified from residual variation
\citep{mathieu2016disentangling,bouchacourt2018multilevel,jha2018cycle,klys2018subspaces,zheng2019label,ilse2020diva,gabbay2020lord}.
These works establish that explicit inductive biases can improve
recombination; they also make a plain VAE latent an inappropriate surrogate
for a declared class-invariant style code. Most directly,
\citet{ridgeway2018leakage} introduce leakage filtering, probe class from a
style posterior, and evaluate content--style recombination on MNIST and more
complex domains. Our novelty is therefore neither the first observation that
content can leak into style nor the algebraic chain rule in isolation. It is
the auditing framework that uses the decomposition to separate four distinct
failure sources, connects finite-sample latent diagnostics to the claim
practitioners make from marginal matching, and validates at decoder level
whether detected leakage changes recombination and sampling.

Image-translation systems operationalize the same recombination requirement:
MUNIT~\citep{huang2018munit} and DRIT~\citep{lee2018drit} combine content with
sampled style, while DMIT~\citep{yu2019dmit} exposes the generator to random
cross-domain combinations during training. Their notion of domain-specific
style is task-dependent---the domain may itself be the conditioning
attribute---but their training strategies highlight the same train--sample
support issue. Recent methods add other inductive biases: V3 exploits
variance--invariance patterns across domains~\citep{wu2025variance}, while
SCFlow learns invertible merging from combinatorial style--content
coverage~\citep{ma2025scflow}. These methods address how to learn a
separation. Reference-guided diffusion work also calls unwanted reference
content carried by style features ``content leakage'' and suppresses it by
masking~\citep{zhu2025less}; our leakage is instead label information in a
learned style latent. We ask when such a latent can be sampled independently.

Several invariance methods directly target $\zs\not\perp y$. The variational
fair autoencoder~\citep{louizos2015variational} matches nuisance-conditional
posteriors to a common target; the per-class style MMD in
Section~\ref{sec:method} is a direct instantiation of that idea, not a new
mechanism. Gradient reversal~\citep{ganin2015unsupervised}, Fader
Networks~\citep{lample2017fader}, and information-theoretic invariance
\citep{moyer2018invariant} remove specified information through different
objectives. They primarily optimize representation invariance; our
decomposition asks the additional question of whether removing
$\zs\not\perp y$ suffices for the independent sampler. It does not when the
other three terms remain.

Deep generative
clustering (DEC~\citep{xie2015unsupervised},
IDEC~\citep{guo2017improved}, VaDE~\citep{jiang2016variational}) evaluates
via ACC/NMI/ARI on the cluster variable, a protocol that cannot see this
failure because it never inspects $\zs$. Our case study makes the point
concrete: $99.15\%$ clustering accuracy on the very checkpoint whose
class-conditional generation reaches $16\%$ external Gen-ACC. For the semantic
variable we use a Spherical Cauchy prior with a M\"obius
reparameterization~\citep{sablica2025hyperspherical}, a heavier-tailed
alternative to the von Mises-Fisher posterior of
S-VAE~\citep{davidson2018hyperspherical}; this is an implementation choice,
not a claim about the leakage, which is defined independently of how $\zc$ is
parameterized.

\section{Factorized Sampling and Its Mismatch}
\label{sec:theory}

Let $\mathcal D=\{(x_i,y_i)\}$ have $K$ classes. A factorized model encodes
each input into a semantic variable $\zc$ and a style variable $\zs$, and
regularizes the latter by
\begin{equation}
\mathcal L_{\mathrm{style}} = D\bigl(q(\zs),\,\mathcal N(0,I)\bigr),
\label{eq:style}
\end{equation}
for some aggregate divergence $D$, with $q(\zs)=\mathbb E_y[q(\zs\mid y)]$.
Write $p(\zs):=\mathcal N(0,I)$ for the style prior. Together with a
class-conditional semantic prior $p(\zc\mid y{=}k)$ these define the
\emph{factorized sampler}: draw $\zc\sim p(\zc\mid y{=}k)$ and $\zs\sim
p(\zs)$ independently, which is the naive class-conditional sampling
procedure in standard use.

\subsection{Marginal matching certifies nothing}

\begin{proposition}[No certificate]
\label{prop:nocert}
Let $\zs\sim\mathcal N(0,I)$ on $\R^{d_s}$ and let $\{A_k\}_{k=1}^K$ be any
measurable partition with $P(\zs\in A_k)=1/K$. Define $y=k \iff \zs\in A_k$.
Then $q(\zs)=\mathcal N(0,I)$ exactly, so $D(q(\zs),\mathcal N(0,I))=0$ for
\emph{every} divergence $D$; and $I(\zs;y)=H(y)=\log K$, the maximum
possible.
\end{proposition}

The proof is immediate: $y$ is defined post hoc as a measurable function of
$\zs$, so the law of $\zs$ is untouched, while $H(y\mid\zs)=0$ makes the
mutual information maximal. The construction is deliberately adversarial and
we do not claim trained decoders produce anything like it. The point is
narrow and worst-case: no divergence computed on $q(\zs)$ alone, MMD or
otherwise, can rule out $\zs$ being maximally class-informative, however the
dependence happens to be shaped.

\subsection{What valid factorized sampling requires}

Proposition~\ref{prop:nocert} is the extremal case of term (2) below. The
following decomposition covers any encoder, not just that construction.

\begin{theorem}[Factorized-sampling mismatch]
\label{thm:decomp}
Let $y$ be uniform on $\{1,\dots,K\}$ and define
$M_{\mathrm{fact}} := \mathbb E_y[\KL(q(\zc,\zs\mid y)\,\|\,p(\zc\mid
y)\,p(\zs))]$. Then
\begin{align*}
M_{\mathrm{fact}} =\;& \underbrace{I_q(\zc;\zs\mid y)}_{\text{(4) within-class dep.}}
 + \underbrace{\mathbb E_y[\KL(q(\zc\mid y)\|p(\zc\mid y))]}_{\text{(3) semantic-prior mismatch}}\\
 &+ \underbrace{I_q(\zs;y)}_{\text{(2) style leakage}}
 + \underbrace{\KL(q(\zs)\|p(\zs))}_{\text{(1) style-prior mismatch}},
\end{align*}
and all four terms are non-negative.
\end{theorem}

The proof is a chain-rule expansion applied twice, given in the
supplementary material. Two consequences matter.

\begin{corollary}[Sampling validity]
\label{cor:valid}
$q(\zc,\zs\mid y{=}k)=p(\zc\mid y{=}k)\,p(\zs)$ for a.e.\ $k$ if and only if
$M_{\mathrm{fact}}=0$, i.e.\ iff all four terms vanish. Class-invariant style
(term 2) is thus necessary but not sufficient; it must be accompanied by a
matched style prior, a matched semantic prior, and within-class independence
of $\zc$ and $\zs$.
\end{corollary}

\begin{corollary}[Decoder pushforward]
\label{cor:push}
For a decoder $\mathrm{Dec}$ with pushforward $\mathrm{Dec}_\#$,
$\mathbb E_y[\KL(\mathrm{Dec}_\# q(\cdot\mid y)\|\mathrm{Dec}_\#[p(\zc\mid
y)p(\zs)])] \le M_{\mathrm{fact}}$ by the data-processing inequality.
\end{corollary}

Corollary~\ref{cor:push} runs one way. A small $M_{\mathrm{fact}}$ suffices
for the fixed decoder to make posterior-decoded and sampler-decoded
distributions close, but it is not by itself a certificate that decoded
samples match the data distribution. Conversely, a large value does not force
a visible failure, since a decoder can be insensitive to the mismatched
directions and map distinct latent distributions to nearly the same images.
Whether a trained decoder actually conditions on the leaked structure is an
empirical question that
the decomposition leaves open, which is why the intervention in
Section~\ref{sec:swap} is needed rather than redundant.

\subsection{Diagnostics}
\label{sec:diag}

Each term gets a finite-sample proxy. We measure the proxies, not the
information quantities, and never estimate $M_{\mathrm{fact}}$ as a whole.
\textbf{Term 1} is proxied by global MMD, $\MMD^2(q(\zs),\mathcal N(0,I))$.
\textbf{Term 2} by inter-class style separation
$\Dinter=\binom{K}{2}^{-1}\sum_{j<k}\|\bar\mus^{(j)}-\bar\mus^{(k)}\|_2$ and
by linear-probe accuracy $\mathrm{LP}(\zs\to y)$. $\Dinter$ is a
mean-separation statistic and thus a convenient symptom, not what
Proposition~\ref{prop:nocert} is about: that proposition concerns dependence
in general and holds even when class means coincide. An HSIC-based proxy
would also apply here, but our fixed-bandwidth estimator saturates at $\zs$'s
empirical scale, returning a near-constant value across checkpoints whose
$\Dinter$ and LP vary widely, so we do not report it. \textbf{Term 3} is
already computed each run as $\mathcal L_{\mathrm{class}}$. \textbf{Term 4}
has no existing proxy; we compare real within-class pairs against
independently recombined ones,
\begin{equation}
\begin{split}
\mathrm{JointMMD} = \tfrac1K\textstyle\sum_k \MMD^2\bigl(&\{(\zc^i,\zs^i)\}_{y_i=k},\\
&\{(\zc^i,\zs^{\pi(i)})\}_{y_i=k}\bigr),
\end{split}
\label{eq:jmmd}
\end{equation}
where $\pi$ permutes style codes within class $k$, breaking the within-class
coupling while leaving both per-class marginals, and hence terms (1)--(3),
untouched.

\section{Method and Case Study}
\label{sec:method}

The standard style regularizer, Eq.~\ref{eq:style}, matches only the
marginal. The remedy we study instead matches each class conditional:
\begin{equation}
\mathcal L_{\text{style-cls}} = \tfrac1K\textstyle\sum_{k=1}^{K}
\MMD_{\mathrm{RBF}}\bigl(\{\zs^i:y_i{=}k\},\{\varepsilon_j\}\bigr),
\label{eq:stylecls}
\end{equation}
$\varepsilon_j\sim\mathcal N(0,I)$. Eq.~\ref{eq:style} has no gradient
incentive to reduce inter-class separation, since the mixture can match
$\mathcal N(0,I)$ while every conditional stays class-structured;
Eq.~\ref{eq:stylecls} penalizes exactly that. Nothing about it is specific to
our architecture: it needs a labeled batch and a style latent. In \ours{},
the semantic MMD and auxiliary classifier already consume the same labels,
so adding Eq.~\ref{eq:stylecls} does not change the supervision regime or
annotation budget. Applied during training to an otherwise unsupervised
model, it would require labels or pseudo-labels; this is distinct from the
diagnostics of Section~\ref{sec:diag}, which need labels only on the
evaluation set and apply post hoc to any trained model.

\ours{} instantiates the setup concretely: a hyperspherical semantic variable
$\zc\in\mathbb S^{d_c-1}$ aligned to class-conditional Spherical Cauchy
priors, and a Euclidean style variable $\zs\in\R^{d_s}$ regularized by
Eqs.~\ref{eq:style}--\ref{eq:stylecls}, with a shared ResNet-18 trunk feeding
two heads and a residual upsampling decoder ($d_c{=}64$, $d_s{=}128$). The
objective adds a supervised semantic MMD, an aggregated semantic MMD, and an
auxiliary classifier on $\muc$ that prevents semantic collapse; these shape
$\zc$ and are not part of the leakage remedy. Coefficients ramp over a
four-phase 300-epoch curriculum in which per-class style regularization
begins in phase B, after a 50-epoch reconstruction-only phase A during which
the decoder does see real $(\zc,\zs)$ pairs unregularized. Section
\ref{sec:exp-robust} tests whether removing that warmup matters; it does not.
Full details are in the supplementary material.

At generation we either sample naively, $\zc\sim p(\zc\mid y{=}k)$ and
$\zs\sim\mathcal N(0,I)$, or apply a post-hoc class-conditional style prior:
estimate $\bar\mus^{(k)}$ and $\bar\sigma_s^{2(k)}$ from encoded training
data and draw $\zs\sim\mathcal N(\bar\mus^{(k)},\tau^2\mathrm{diag}(\bar\sigma_s^{2(k)}))$
with $\tau{=}0.25$. The second needs no retraining, only a forward pass.

\section{Experiments}

CIFAR-10~\citep{krizhevsky2009learning} is the primary
clustering/generation benchmark; MNIST~\citep{lecun1998gradient} and
Fashion-MNIST~\citep{xiao2017fashion} are used for diagnostics. We report
clustering (ACC/NMI/ARI), reconstruction (SSIM/LPIPS~\citep{zhang2018unreasonable}),
sample quality (FID~\citep{heusel2017gans}, 10k vs 10k), and generated-class
accuracy (Gen-ACC), the fraction of generated images assigned to the sampled
class by an independently trained external classifier.

\begin{table}[t]
\centering
\small
\setlength{\tabcolsep}{5pt}
\caption{CIFAR-10 against baselines matched in backbone (ResNet-18) and
latent dimension ($d{=}192$), grouped by how the label is used. \ours{} is
3-seed mean. The fair clustering comparison is against the label-guided
group, which like \ours{} pushes the label into the latent.}
\label{tab:cifar-main}
\begin{tabular}{lrrr}
\toprule
Method & ACC & NMI & FID \\
\midrule
\textbf{\ours{}} & \textbf{0.809} & \textbf{0.650} & \textbf{83.0} \\
\midrule
\multicolumn{4}{l}{\emph{Label-guided (label shapes the latent)}} \\
AEWithCE & 0.665 & 0.611 & 144.3 \\
AEWithSupCon & 0.541 & 0.599 & 275.5 \\
AEWithTriplet & 0.506 & 0.502 & 281.4 \\
AEWithCenterLoss & 0.243 & 0.122 & 235.8 \\
\multicolumn{4}{l}{\emph{Conditional-generative (label $\to$ decoder)}} \\
ConditionalVAE & 0.133 & 0.024 & 396.4 \\
ConditionalWAE-MMD & 0.102 & 0.008 & 170.3 \\
GaussianClassPriorWAE & 0.103 & 0.008 & 174.7 \\
\multicolumn{4}{l}{\emph{Unsupervised}} \\
ResNetAE & 0.220 & 0.102 & 121.0 \\
\bottomrule
\end{tabular}
\end{table}

Before using \ours{} to study leakage we check it is not degenerate, since a
degenerate model could leak trivially. Table~\ref{tab:cifar-main} compares it
against seven baselines matched in backbone and latent dimension. Against the
label-guided autoencoders, the fair comparison, \ours{} ($80.9\%$) exceeds the
best ($66.5\%$) and has the lowest FID in the table. The
conditional-generative baselines cluster near chance because their label
enters only at the decoder, leaving the encoder latent unshaped; they are
included as the closest generative analogues, not as a clustering
comparison. This is evidence of competitiveness, not a controlled
superiority claim: the baselines are single-seed, \ours{} uses both
latent-shaping and a generative prior where each baseline uses one, and we
did not equalize per-method tuning.

\paragraph{Evaluation independence.}
All generation-side metrics in the main paper are scored directly from
generated pixels by a classifier trained only on real images and independent
of the model under test: a 4-layer CNN for MNIST ($99.4\%$ real-test
accuracy) and a WideResNet-28-10~\citep{zagoruyko2016wide} for CIFAR-10
($94.1\%$). Under this fixed
protocol, MNIST naive Gen-ACC is $0.16$, the class-conditional prior reaches
$0.97$, and latent-swap style-following is $96.5\%$. On CIFAR-10, the
class-conditional diagonal prior at $\tau{=}0.25$ reaches $0.41$ and the
empirical style bank reaches $0.88$. Internal-classifier scores are reported
only in the supplementary material as a robustness comparison; they are not
mixed with the primary results below.

\subsection{Marginal metrics hide the leakage}
\label{sec:exp-hide}

\begin{table}[t]
\centering
\small
\setlength{\tabcolsep}{4pt}
\caption{Conditional style leakage across a case-study model and four
reference latent baselines on MNIST.
Global MMD is near zero everywhere; $\Dinter$ and LP are not. Gen-ACC is
scored by the external CNN and uses naive $\zs\sim\mathcal N(0,I)$.
The four reference families share one architecture and budget; \ours{} uses
its native configuration, so conclusions are within-row rather than ranked
across rows.}
\label{tab:cross}
\begin{tabular}{lcccc}
\toprule
Model & MMD & $\Dinter$ & LP & Gen-ACC \\
\midrule
VAE           & 0.0002 & 2.86 & 87.4\% & N/A \\
WAE-MMD       & 0.0576 & 9.38 & 93.0\% & N/A \\
$\beta$-TCVAE & 0.0002 & 2.30 & 73.9\% & N/A \\
FactorVAE     & 0.0009 & 2.27 & 75.9\% & N/A \\
\ours{}       & 0.0013 & 6.27 & 100.0\% & 0.16 \\
\midrule
\ours{}\,$+$\,per-class & 0.0015 & 1.22 & 42.6\% &
0.66 \\
\bottomrule
\end{tabular}
\end{table}

Table~\ref{tab:cross} applies the diagnostic to four unsupervised baselines
and \ours{} on MNIST. The baselines are purpose-built for this comparison
and share the same encoder family, latent dimension, optimizer, and
100-epoch budget with one another. \ours{} is evaluated in its native
configuration. Accordingly, the table is an audit rather than a leaderboard:
for every row individually, global MMD is decoupled from $\Dinter$ and LP,
and the argument does not depend on ranking leakage severity across model
families.

Every baseline shows LP far above chance despite near-zero global MMD.
WAE-MMD has the largest global MMD ($0.0576$, some $30$--$300\times$ the
others), suggesting its own regularizer had not converged, and correspondingly
the largest baseline $\Dinter$ and LP. More telling are $\beta$-TCVAE and
FactorVAE: their objectives explicitly penalize latent dependence, and they
do achieve the smallest $\Dinter$ of the five, yet LP stays at $74$--$76\%$.
Penalizing total correlation among latent \emph{dimensions} reduces, but does
not remove, dependence between the latent and the \emph{label}. These are
different quantities.

\begin{figure}[t]
\centering
\includegraphics[width=\columnwidth]{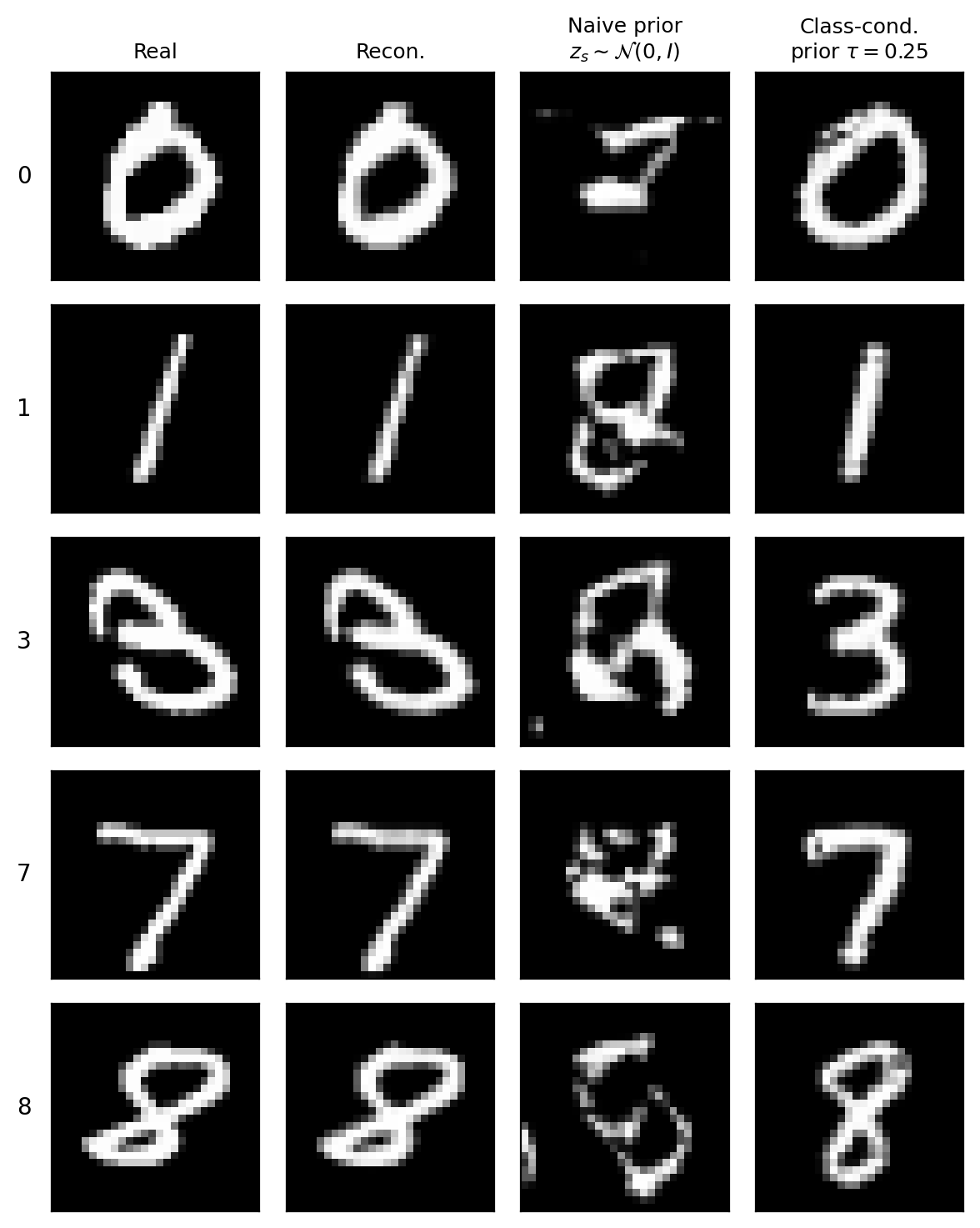}
\caption{Real image, posterior reconstruction, naive-prior sample, and
class-conditional-prior sample, for five digits. Reconstructions are
indistinguishable from the real image (SSIM $0.983$); naive-prior samples are
legible digits but frequently the \emph{wrong} one; the class-conditional
prior restores the intended identity.}
\label{fig:recongen}
\end{figure}

\ours{} without the remedy is the most extreme case ($\Dinter=6.27$,
LP$=100\%$), and the consequence is concrete rather than statistical. On the
same checkpoint it reaches $99.15\%$ clustering accuracy and SSIM $0.983$,
yet class-conditional generation under naive sampling reaches only $16\%$
external Gen-ACC. The failure is also structured rather than random: probability mass
collapses onto a handful of attractor digits regardless of the class
requested, which is what one expects if the decoder resolves an unfamiliar
$(\zc,\zs)$ pair by trusting whichever code it learned to read identity from.
A 2D t-SNE of $\zs$ makes the leakage visually obvious only for the two
largest-$\Dinter$ rows; for VAE, $\beta$-TCVAE, and FactorVAE it is
statistically real but invisible in projection, which is a reason to prefer
the quantitative diagnostic over a plot.

\subsection{The decoder reads identity off the style code}
\label{sec:swap}

A probe shows $\zs$ \emph{contains} class information; it does not show the
decoder \emph{uses} it. For every ordered class pair $(a,b)$ we take $\muc$
from a real test image of class $a$ and $\mus$ from an independent real image
of class $b$, decode, and classify. This uses only real, individually-encoded
outputs rather than prior draws, so a failure to recover class $a$ cannot be
blamed on an out-of-support prior sample, though the pairing itself may lie
off the joint support the decoder trained on.

\begin{figure}[t]
\centering
\includegraphics[width=\columnwidth]{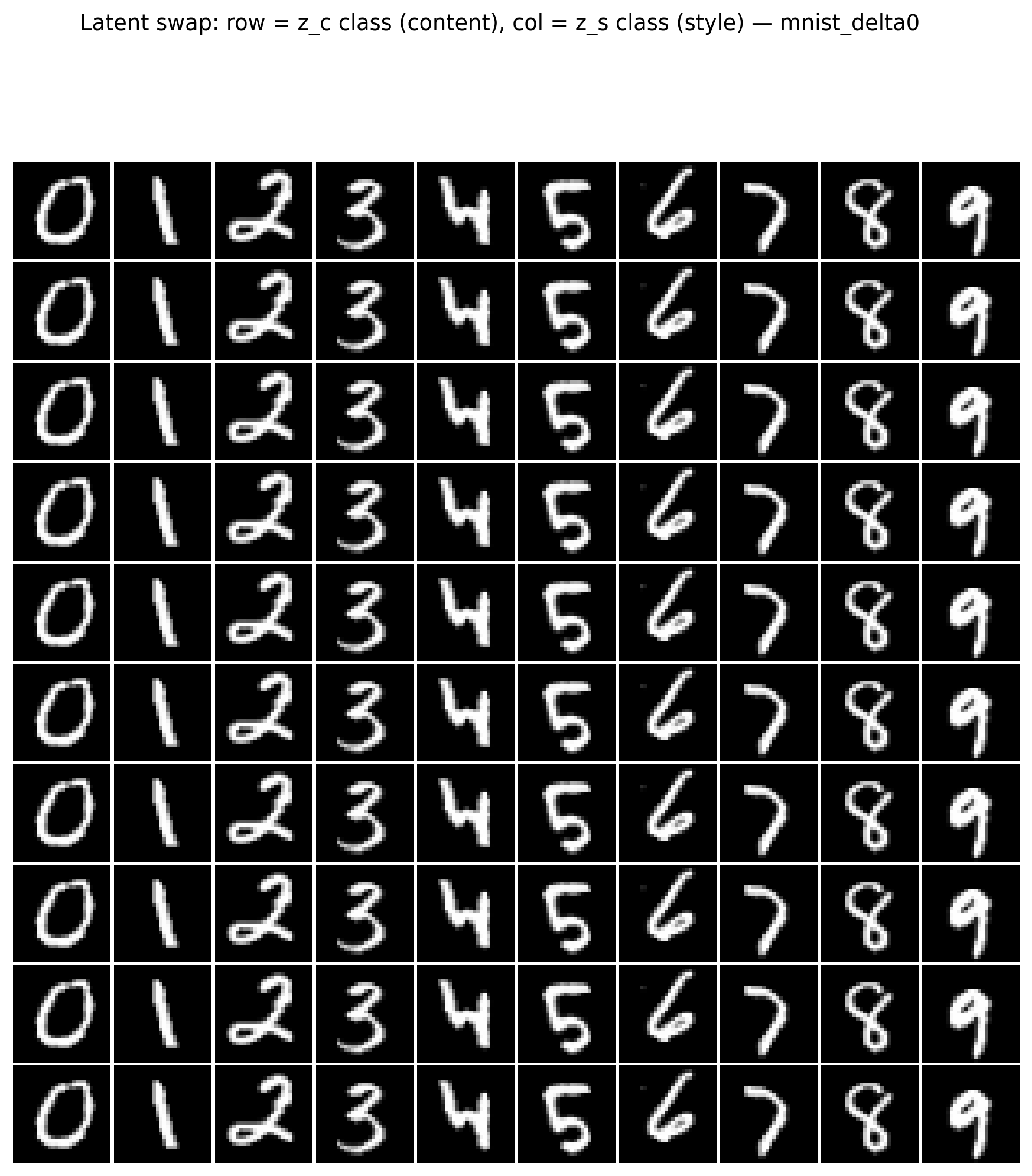}
\caption{MNIST latent swap at $\delta{=}0$: rows donate $\zc$ (content) and
columns donate $\zs$ (style). The external CNN assigns the output to the
style-donor class in $96.5\%$ of off-diagonal swaps. Internal-classifier
heatmaps for both $\delta$ settings are retained only as a supplementary
robustness comparison.}
\label{fig:swapheat}
\end{figure}

At $\delta{=}0$ on MNIST, external style-following is $96.5\%$
(Figure~\ref{fig:swapheat}): the decoder reads identity predominantly from
$\zs$. Every row of the swap grid (fixed $\zc$) looks nearly identical; the
column, the $\zs$ donor, determines the digit. The per-class remedy shifts
the qualitative dependence toward content on both datasets. Exact rates
under the model's internal classifier are provided only in the supplementary
robustness comparison and are not used as primary evidence here.

Per-class conditional MMD confirms the same structure behind $\Dinter$'s
single scalar: at $\delta{=}0$, mean per-class $\MMD^2(q(\zs\mid
y{=}k),\mathcal N(0,I))$ is $21\times$ the global MMD on MNIST ($0.0186$ vs
$0.00089$) and $15\times$ on CIFAR-10. Every one of the ten per-class values
sits far above the global figure the marginal regularizer actually optimizes.
At $\delta{=}1$ the gap narrows but does not close ($4.4\times$; $5.3\times$).

\paragraph{Is this an off-support artifact?}
The obvious objection is that cross-class pairs simply leave the training
joint's support. Grading the intervention by support answers it. Same-class
cross-image pairs sit at $1.2\times$ the same-image kNN distance to the
training joint and preserve identity, whereas cross-class pairs sit at
$2.1\times$ and largely lose the content-donor identity. Interpolating
$\zs$ between a same-class and a cross-class donor hands identity over
monotonically, crossing at $\alpha\approx0.38$, well before the pair reaches
cross-class support distance. Style-dominance turns on as soon as $\zs$
carries a competing class signal, not only once the pair leaves the support.
Cross-class pairs are measurably farther out, so the objection is not fully
dissolved, but the effect does not reduce to that distance.

\subsection{What the remedies do and do not fix}
\label{sec:exp-remedy}

Per-class style MMD works in the intended direction without finishing the
job. It cuts $\Dinter$ by $81\%$ ($6.27\to1.22$) and LP by $57.4$ points
($100\%\to42.6\%$), still far above the $10\%$ floor and higher than three of
the four unsupervised baselines. Theorem~\ref{thm:decomp} explains why this
term-2 remedy is necessarily partial: Eq.~\ref{eq:stylecls} drives terms
(1)--(2) down by construction but places no pressure on term (4), the
within-class dependence a decoder can still exploit once every
$q(\zs\mid y{=}k)$ individually matches the prior.

\begin{table}[t]
\centering
\small
\setlength{\tabcolsep}{3.5pt}
\caption{Representation-level remedies on MNIST at matched backbone,
dimensions, and training budget. Generation scores from the model's internal
classifier are moved to the supplementary robustness comparison. Single-seed
point estimates are reported.}
\label{tab:invariance}
\scriptsize
\setlength{\tabcolsep}{2.5pt}
\begin{tabular}{lrrr}
\toprule
Remedy & LP & $\Dinter$ & JointMMD \\
\midrule
None ($\delta{=}0$) & 100.0 & 6.27 & .0037 \\
\midrule
Per-cls.\ MMD ($\delta{=}1$) & 42.6 & 1.22 & .0043 \\
{VFAE-style cond.} & 46 & 1.25 & .0042 \\
{HSIC penalty} & 37 & 1.08 & .0041 \\
{Gradient reversal} & 21 & 0.82 & .0044 \\
\midrule
{$+$ joint term-4} & 40 & 1.15 & .0016 \\
\bottomrule
\end{tabular}
\end{table}

\begin{table}[t]
\centering
\small
\setlength{\tabcolsep}{4pt}
\caption{Six-point sweep of the per-class style MMD weight $\delta$.
Single-seed point estimates are reported. $\Dinter$ decreases steadily on
both datasets, while ACC is non-monotonic and has one pronounced dip per
dataset.}
\label{tab:tradeoff}
\scriptsize
\setlength{\tabcolsep}{2.5pt}
\begin{tabular}{lrrrr}
\toprule
& \multicolumn{2}{c}{MNIST} & \multicolumn{2}{c}{CIFAR-10} \\
$\delta$ & ACC & $\Dinter$ & ACC & $\Dinter$ \\
\midrule
0     & .992 & 6.21 & .813 & 3.74 \\
0.03  & .994 & 3.27 & .811 & 2.48 \\
0.1   & .990 & 2.12 & .810 & 2.03 \\
0.3   & .994 & 1.55 & .715 & 1.59 \\
1     & .859 & 1.22 & .800 & 1.05 \\
3     & .993 & 1.18 & .799 & 0.95 \\
\bottomrule
\end{tabular}
\end{table}

Table~\ref{tab:tradeoff} sweeps six values. $\Dinter$ decreases steadily on
both datasets, so the leakage-reduction mechanism behaves as expected across
the range rather than only at the endpoints. Naive FID broadly improves but
is not monotonic, and clustering ACC has a pronounced single-seed dip at
$\delta{=}1$ on MNIST and $\delta{=}0.3$ on CIFAR-10. With one seed per
point, we cannot attribute those dips to $\delta$ rather than initialization
and make no significance claim across sweep settings.

Because per-class MMD only reduces term 2 to $42.6\%$ LP, it cannot separate
``term 2 is still too large'' from ``term 2 is not the whole story''.
Table~\ref{tab:invariance} therefore compares four remedies that attack term
2 through different mechanisms, plus one that also targets term 4.
{Gradient reversal is much the strongest term-2 remedy, driving LP to
$21\%$, within $11$ points of chance and less than half what per-class MMD
achieves. Yet every term-2-only intervention leaves JointMMD in the narrow
$0.0041$--$0.0044$ range. Targeting term 4 directly cuts JointMMD by
$2.7\times$, the only intervention that moves this proxy substantially,
while its LP remains higher than gradient reversal's. Thus reducing
$I(\zs;y)$ does not by itself reduce within-class $\zc$--$\zs$ dependence,
exactly the distinction made by Corollary~\ref{cor:valid}. Internal-evaluator
generation scores for these checkpoints are retained only as a robustness
comparison in the supplementary material.}

We calibrate the term-4 proxy against a within-class double-permutation null
using 200 permutations. {Every checkpoint sits $4$--$5\times$ above its null
($p{=}0.005$), so term
4 is genuinely non-zero, while the $\delta{=}0\to1$ change is not significant
($p{=}0.31$): per-class style MMD leaves term 4 statistically untouched even
as it cuts $\Dinter$ and LP sharply. A conditional-HSIC cross-check agrees.}
One comparison we avoid: MNIST and CIFAR-10 JointMMD values are close, but
even after standardization, equal MMD on two different latent distributions
does not imply equal dependence, so we draw no cross-dataset conclusion from
it.

\subsection{Transfer and robustness}
\label{sec:exp-robust}

\begin{table}[t]
\centering
\small
\setlength{\tabcolsep}{3.5pt}
\caption{Single-variable perturbations from the $\delta{=}0$ checkpoint,
each removing one candidate confound, both datasets. Single-seed point
estimates are reported.}
\label{tab:audit}
\scriptsize
\setlength{\tabcolsep}{2.5pt}
\begin{tabular}{lrrrr}
\toprule
& \multicolumn{2}{c}{MNIST} & \multicolumn{2}{c}{CIFAR-10} \\
Perturbation & $\Dinter$ & LP & $\Dinter$ & LP \\
\midrule
None (baseline) & 6.27 & 100.0 & 3.80 & 86.9 \\
Balanced dims & 4.79 & 99.9 & 2.11 & 52.4 \\
Smaller style & 3.28 & 92.6 & 1.68 & 44.0 \\
No warmup & 5.53 & 100.0 & 3.78 & 88.3 \\
Gaussian prior & 5.35 & 99.6 & 3.82 & 74.5 \\
vMF prior & 5.69 & 100.0 & 3.03 & 83.0 \\
No classifier & 5.09 & 97.1 & 3.15 & 67.1 \\
\bottomrule
\end{tabular}
\end{table}

\begin{figure}[t]
\centering
\includegraphics[width=\columnwidth]{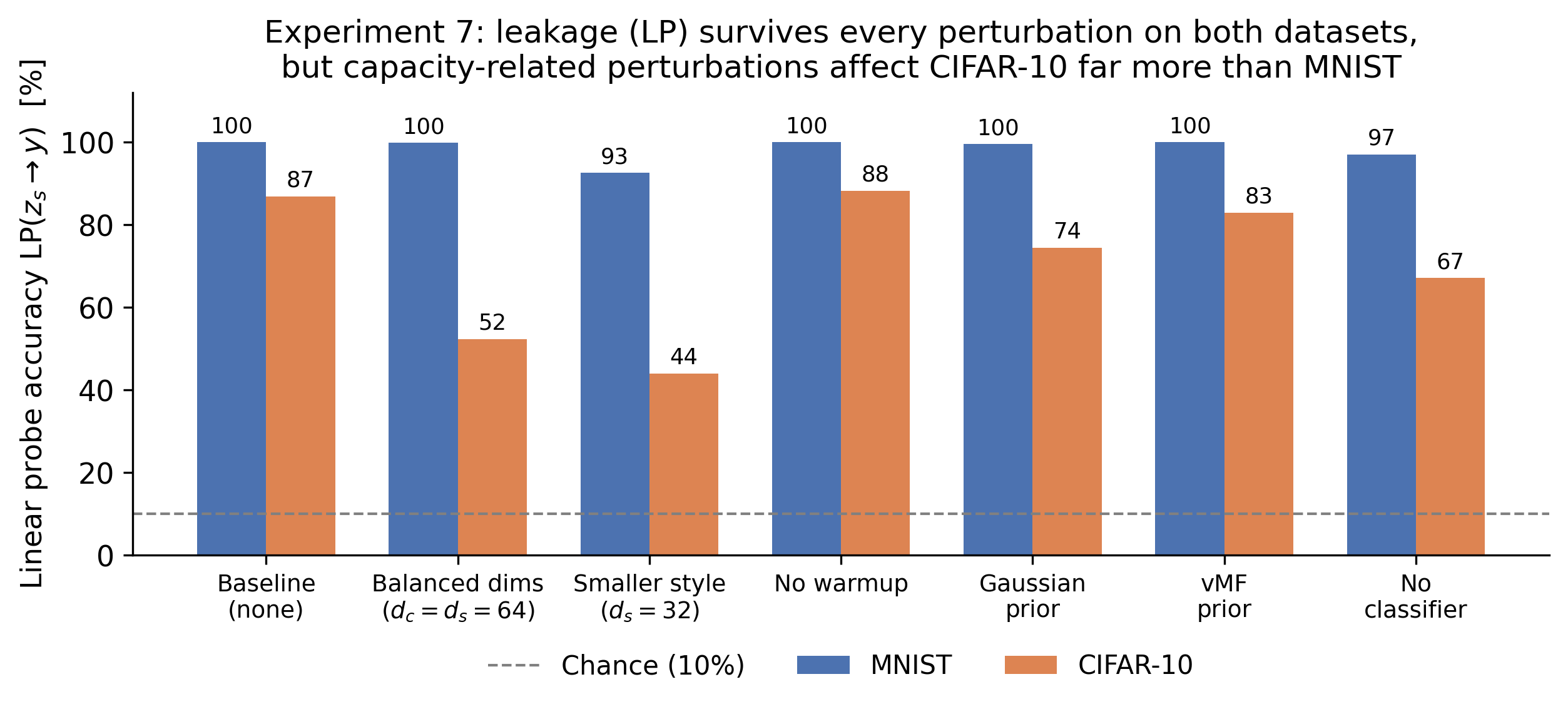}
\caption{Linear-probe recovery of the label from $\zs$ for the baseline and
six perturbations. Leakage survives all of them on both datasets, but the
two capacity perturbations move CIFAR-10 far more than they move MNIST,
while no-warmup and the vMF prior barely move either.}
\label{fig:audit}
\end{figure}

Several design choices in \ours{} could plausibly be the real explanation for
why $\zs$ absorbs so much class information: the style latent being larger
than the semantic one, the reconstruction-only warmup, the prior geometry, or
the auxiliary classifier. Table~\ref{tab:audit} and Figure~\ref{fig:audit}
perturb each independently on both datasets. LP never drops below $92.6\%$ on MNIST or $44.0\%$ on
CIFAR-10, and neither approaches chance. On MNIST every perturbation is mild,
costing at most $7.4$ points; on CIFAR-10 the same six span a $44$-point
range. Smaller style capacity is the largest effect on both ($7.4$ and $42.9$
points), followed by balanced dims, but capacity is not uniquely special:
removing the classifier costs $19.8$ points on CIFAR-10 and the Gaussian
prior $12.4$, while no-warmup and the vMF prior barely move it.

Two readings follow. Capacity is a real, dataset-dependent contributor,
consistent with severity tracking how much class structure a dataset forces
into $\zs$, and a paper that checked only MNIST would have badly understated
how much one hyperparameter can move this number. We also flag a confound we
do not disentangle: the perturbations that reduce LP most also reduce
representational capacity, and removing the classifier collapses clustering
from $99\%$ to $55.4\%$, so part of each drop may reflect a weaker
representation rather than genuine invariance. Even so, no single tested
perturbation drives LP near chance, so none of these choices \emph{eliminates}
the leakage, even though capacity clearly modulates it.

Leakage severity also follows the expected dataset ordering. {Measuring
intra-class diversity independently of our model, as mean within-class
pairwise cosine distance in a frozen DINOv2 space, confirms the ordering with
clear separation: MNIST $0.33$, Fashion-MNIST $0.47$, CIFAR-10 $0.66$.} Both
$\Dinter$ and LP decrease along it (MNIST $6.27$/$100\%$, Fashion-MNIST
$4.86$/$97.7\%$, CIFAR-10 $3.80$/$86.9\%$). We do not use Gen-ACC as a
cross-dataset proxy for leakage severity: unlike $\Dinter$ and LP, it is
measured after decoding and therefore also inherits dataset difficulty,
sample quality, and external-classifier error.

\paragraph{The repair does not transfer.}

\begin{table}[t]
\centering
\scriptsize
\setlength{\tabcolsep}{2.5pt}
\caption{Generation under $\zs$ sampling strategies on the $\delta{=}0$
checkpoints ($\zc$ always from the class prior). Gen-ACC uses the fixed
external CNN on MNIST and WideResNet-28-10 on CIFAR-10.}
\label{tab:sampling}
\begin{tabular}{lrrrr}
\toprule
& \multicolumn{2}{c}{MNIST} & \multicolumn{2}{c}{CIFAR-10} \\
Style sampling $\zs$ & Gen-ACC & Div. & Gen-ACC & Div. \\
\midrule
Global $\mathcal N(0,I)$ & 0.16 & 0.151 &
 0.09 & 0.187 \\
Class diag., $\tau{=}0.25$ & \textbf{0.97} & 0.135 & 0.41 & 0.101 \\
Empirical style bank & 0.96 & 0.182 &
\textbf{0.88} & \textbf{0.241} \\
\bottomrule
\end{tabular}
\end{table}

On MNIST, the external CNN raises Gen-ACC from $0.16$ under global Gaussian
sampling to $0.97$ under the class-conditional diagonal prior at
$\tau{=}0.25$ (Table~\ref{tab:sampling}). On CIFAR-10, the same parametric
recipe reaches only $0.41$ under the external WideResNet. The empirical
style bank reaches $0.88$ and also has higher measured diversity ($0.241$
versus $0.101$), indicating that matching only a unimodal diagonal Gaussian
misses important shape in $q(\zs\mid y{=}k)$. Controls for mean shift,
variance shrinkage, and shuffled labels show the same qualitative mechanism
under the internal evaluator and are reported only as a supplementary
robustness comparison. The diagnostic therefore transfers across datasets
more cleanly than the parametric repair does.

\section{Discussion}

CIFAR-10's higher intra-class diversity weakens conditional style dependence,
and $\Dinter$/LP confirm this cleanly, but the practical remedy does not
inherit that cleanliness. CIFAR-10 samples stay visibly blurry (FID
$\sim83$--$84$), and we avoid calling naive output ``visually plausible''; at
this FID the accurate description is recognizable but not sharp. Global
aggregate matching is standard in WAEs and disentanglement models, and our
results show it is insufficient whenever the style variable must be
class-invariant for generation. Per-class style MMD is best read as a
supervised analogue of total-correlation penalties, aimed at the single
dependency $\zs\not\perp y$ rather than at dependence among latent dimensions
in general. Finally, the MNIST checkpoint reaches SSIM $0.983$ yet FID
$73.99$ under naive sampling: reconstruction measures fidelity on the
posterior manifold, not the prior, and evaluation of generative clustering
should report both, since the two can diverge substantially.

\paragraph{Limitations.}
We report no component-wise ablation beyond the $\delta$ axis. The per-class remedy
assumes labels already available to the label-guided training objective and
is not, by itself, an unsupervised disentanglement loss. Gen-ACC is
confounded by dataset-intrinsic difficulty. The CIFAR-10 transfer failure is
measured on one checkpoint per dataset, and we have not tested whether a
richer conditional (a per-class mixture) closes it. The robustness check
perturbs one variable at a time, so joint perturbations, which could compound
given how much larger each effect is on CIFAR-10, remain untested.
{Table~\ref{tab:invariance} is MNIST-only; whether gradient reversal's
term-2 advantage and the joint remedy's term-4 advantage replicate on
CIFAR-10 is untested. Three datasets remain too few to treat the
diversity--leakage relationship as a measured correlation rather than a
consistent ordering.}

\section{Conclusion}

Conditional style leakage is a real failure mode in factorized generative
models: aggregate regularization of a style variable does not prevent it from
becoming class-dependent, because marginal matching is compatible with $\zs$
being maximally informative about the label rather than merely correlated
with it. We placed it inside an exact decomposition of what factorized
sampling requires, measured it across a case-study model and four reference
latent baselines, and showed that a decoder trained this way is measurably
more sensitive to the leaked code than
to the semantic one. No single perturbation of capacity, curriculum, prior
geometry, or supervision eliminates the leakage on either dataset, though
capacity strongly modulates its magnitude on CIFAR-10. {Comparing four
mechanistically different term-2 remedies settles what one remedy could not:
they span $21$--$46\%$ probe recovery yet leave the term-4 proxy unchanged,
whereas the joint remedy cuts that proxy by $2.7\times$ despite a higher
probe score than gradient reversal. Eliminating conditional style leakage is
necessary but not sufficient, exactly as Corollary~\ref{cor:valid} states.}
Under the fixed external protocol, the conditional prior reaches $0.97$
Gen-ACC on MNIST but only $0.41$ on CIFAR-10; an empirical bank reaches
$0.88$. Thus fixing a diagnostic finding on one dataset does not fix the
underlying generative gap on another. Factorized generative models should
verify class-invariance of style variables with conditional diagnostics
rather than marginal statistics, and should report where their remedies do
and do not generalize.

\clearpage

\bibliography{references}

\clearpage
\begin{center}
{\LARGE\bfseries Supplementary Material\\[0.4em]
Marginal Matching Does Not License Factorized Sampling:\\
Auditing Conditional Style Leakage in Factorized Generative Models\par}
\end{center}

\section{A. Proofs}

\subsection{A.1 Proposition 1 (No certificate)}

\begin{proposition}
Let $\zs\sim\mathcal N(0,I)$ on $\R^{d_s}$, and let $\{A_k\}_{k=1}^K$ be any
measurable partition of $\R^{d_s}$ with $P(\zs\in A_k)=1/K$ for each $k$
(such a partition always exists, e.g.\ via level sets of a continuous
statistic of $\zs$). Define $y=k \iff \zs\in A_k$. Then (i) $q(\zs)=\mathcal
N(0,I)$ exactly, so $D(q(\zs),\mathcal N(0,I))=0$ for every divergence $D$;
and (ii) $I(\zs;y)=H(y)=\log K$.
\end{proposition}

\begin{proof}
(i) $y$ is defined post hoc as a measurable function of $\zs$; the
distribution of $\zs$ itself is never altered, so $q(\zs)=\mathcal N(0,I)$
holds exactly, and every divergence to $\mathcal N(0,I)$, aggregate MMD
included, is identically zero. (ii) Since $y$ is a deterministic function of
$\zs$, $H(y\mid\zs)=0$, so $I(\zs;y)=H(y)-H(y\mid\zs)=H(y)$. By construction
$P(y=k)=1/K$ for all $k$, so $H(y)=\log K$, which is also the maximum
possible entropy, and hence the maximum possible mutual information with any
other variable, for a $K$-ary label.
\end{proof}

The construction is deliberately adversarial: $\{A_k\}$ need not resemble
anything a trained decoder produces, and $q(\zs\mid y{=}k)$ here is $\zs$
restricted to $A_k$ and renormalized, not a smooth family with class-specific
means. That is the point. It shows that no divergence computed on $q(\zs)$
alone can rule out $\zs$ being maximally class-informative, regardless of how
the dependence is structured geometrically. It is a worst-case
non-identifiability result, not a claim about typical behavior.

\subsection{A.2 Theorem 1 (Factorized-sampling mismatch)}

\begin{theorem}
Let $y$ be uniform on $\{1,\dots,K\}$, let $q(\zc,\zs\mid y{=}k)$ be the
encoder's joint class conditional, let $p(\zc\mid y{=}k)$ and $p(\zs)$ be the
semantic and style priors, and let $q(\zs)=\mathbb E_y[q(\zs\mid y)]$. Define
\[
M_{\mathrm{fact}} := \mathbb E_y\bigl[\KL(q(\zc,\zs\mid y)\,\|\,p(\zc\mid y)\,p(\zs))\bigr].
\]
Then
\begin{align*}
M_{\mathrm{fact}} =\;& I_q(\zc;\zs\mid y)
 + \mathbb E_y[\KL(q(\zc\mid y)\|p(\zc\mid y))]\\
 &+ I_q(\zs;y) + \KL(q(\zs)\|p(\zs)),
\end{align*}
and all four terms are non-negative.
\end{theorem}

\begin{proof}
Fix $k$ and expand the class-$k$ term by inserting $\pm\log q(\zc\mid
k)\pm\log q(\zs\mid k)$ into the integrand:
\begin{align*}
&\KL\bigl(q(\zc,\zs\mid k)\,\|\,p(\zc\mid k)\,p(\zs)\bigr) \\
&= \mathbb E_{q(\zc,\zs\mid k)}\Bigl[\log\tfrac{q(\zc,\zs\mid k)}{q(\zc\mid k)\,q(\zs\mid k)}\Bigr]\\
&\quad + \mathbb E_{q(\zc\mid k)}\Bigl[\log\tfrac{q(\zc\mid k)}{p(\zc\mid k)}\Bigr]
 + \mathbb E_{q(\zs\mid k)}\Bigl[\log\tfrac{q(\zs\mid k)}{p(\zs)}\Bigr] \\
&= I_q(\zc;\zs\mid y{=}k) + \KL(q(\zc\mid k)\|p(\zc\mid k))\\
&\quad + \KL(q(\zs\mid k)\|p(\zs)),
\end{align*}
where the first term is, by definition, the mutual information between $\zc$
and $\zs$ under $q(\cdot,\cdot\mid y{=}k)$. Taking $\mathbb E_y[\cdot]$ gives
the first two terms of the statement directly, since $\mathbb
E_y[I_q(\zc;\zs\mid y{=}k)]=I_q(\zc;\zs\mid y)$ by definition of conditional
mutual information. For the remaining piece, $\mathbb E_y[\KL(q(\zs\mid
y)\|p(\zs))]$, insert $\pm\log q(\zs)$ and use $q(\zs)=\mathbb E_y[q(\zs\mid
y)]$:
\begin{align*}
\mathbb E_y\bigl[\KL(q(\zs\mid y)\|p(\zs))\bigr]
={}& \underbrace{\mathbb E_y\bigl[\KL(q(\zs\mid y)\|q(\zs))\bigr]}_{=\,I_q(\zs;y)}\\
&+ \KL(q(\zs)\|p(\zs)).
\end{align*}
where the first bracket is exactly $I_q(\zs;y)$: the average KL of a
conditional from its own marginal is, by definition, the mutual information.
Non-negativity of every term follows from non-negativity of KL divergence and
of mutual information (itself a KL divergence).
\end{proof}

\subsection{A.3 Corollaries}

\begin{corollary}[Exact sampling validity]
$q(\zc,\zs\mid y{=}k)=p(\zc\mid y{=}k)\,p(\zs)$ for almost every $k$ if and
only if $M_{\mathrm{fact}}=0$, i.e.\ iff all four terms vanish
simultaneously. In particular, class-invariant style ($I_q(\zs;y)=0$) is
necessary but not sufficient.
\end{corollary}

\begin{proof}
Every term of Theorem 1 is a non-negative expectation of a KL divergence
(terms 2 and 4 are mutual informations, themselves KL divergences), so their
sum $M_{\mathrm{fact}}$ is zero iff each term is zero; and $\mathbb
E_y[\KL(\cdot\|\cdot)]=0$ iff the integrand vanishes for a.e.\ $y$, iff the
two per-class distributions coincide a.e.
\end{proof}

\begin{corollary}[Decoder pushforward bound]
Let $\mathrm{Dec}$ be the decoder and $\mathrm{Dec}_\#\mu$ the law of
$\mathrm{Dec}(\zc,\zs)$ when $(\zc,\zs)\sim\mu$. Then
\[
\mathbb E_y\bigl[\KL(\mathrm{Dec}_\# q(\cdot\mid y)\,\|\,\mathrm{Dec}_\#[p(\zc\mid y)\,p(\zs)])\bigr] \le M_{\mathrm{fact}}.
\]
\end{corollary}

\begin{proof}
The data-processing inequality states that pushing two distributions through
the same Markov kernel (here $\mathrm{Dec}$) cannot increase their KL
divergence; applying it inside $\mathbb E_y[\cdot]$ and invoking Theorem 1
gives the bound.
\end{proof}

This runs one way only. A small $M_{\mathrm{fact}}$ suffices for the fixed
decoder to make posterior-decoded and sampler-decoded class-conditional
distributions close, but it is not by itself a certificate that decoded
samples match the data distribution. Conversely, a large value does not force
a visible failure, because a decoder can be insensitive to the mismatched
latent directions and map distinct latent distributions to nearly identical
image distributions. Whether the trained decoder actually conditions on the
leaked structure that terms (2) and (4) quantify is an empirical question the
decomposition does not settle, which is why the latent-swap intervention in
the main paper is necessary rather than redundant.

\section{B. Architecture Details}

The semantic head produces $(\tilde\muc,r_c)$, giving
$\muc=\tilde\muc/\|\tilde\muc\|_2$ and concentration
$\rho_c=\sigma(r_c)(1-\varepsilon)$; the semantic latent is sampled by the
Spherical Cauchy M\"obius reparameterization
$\eta_c\sim\mathrm{Unif}(\Sph^{d_c-1})$, $\zc=T_{\muc,\rho_c}(\eta_c)$. The
style head produces $(\mus,\log\sigma_s^2)$ and $\zs=\mus+\sigma_s\odot
\varepsilon_s$, $\varepsilon_s\sim\mathcal N(0,I)$.

The ResNet-18 backbone operates on $32\times32$ inputs. It replaces the first $7\times7$
convolution (stride 2) with a $3\times3$ convolution (stride 1, padding 1)
and removes the max-pool layer, yielding a $4\times4$ spatial output with 512
channels; a shared FC layer maps $512\times4\times4{=}8{,}192$ to 256
dimensions with SiLU, from which separate linear heads produce the semantic
and style parameters. The decoder concatenates $[\zc;\zs]$ ($d_c{=}64$,
$d_s{=}128$ in all experiments unless stated) and maps it through three
ResBlockUp stages (bilinear $2\times$ upsample, $1\times1$ projection,
GroupNorm+SiLU+Conv residual block; $512{\to}256{\to}128{\to}64$ channels,
$4\times4\to32\times32$), followed by a $3\times3$ output convolution and
sigmoid.

For each class $k$, \ours{} maintains a center $m_k\in\Sph^{d_c-1}$ with
prior $p(\zc\mid y{=}k)=\SCauchy(m_k,\rho_p)$, $\rho_p{=}0.7$, updated by EMA
from batch class-mean directions once per epoch,
$m_k\leftarrow\mathrm{normalize}(\tau m_k+(1-\tau)\bar\muc^{(k)})$,
$\tau{=}0.95$. The full training objective is
\begin{align*}
  \mathcal{L} &= \mathcal{L}_{\mathrm{rec}} + \alpha(t)\mathcal{L}_{\mathrm{class}}
   + \beta(t)\mathcal{L}_{\mathrm{agg}} + \gamma(t)\mathcal{L}_{\mathrm{style}}\\
  &\quad + \delta(t)\mathcal{L}_{\mathrm{style\text{-}cls}} + \eta(t)\mathcal{L}_{\mathrm{cls}},
\end{align*}
with
$\mathcal{L}_{\mathrm{rec}}=\lambda_1\|x-\hat x\|_1+\lambda_{\mathrm{LPIPS}}\mathrm{LPIPS}(x,\hat x)$,
$\mathcal{L}_{\mathrm{class}}=\frac1K\sum_k\MMD^2(Q_k,P_k)$ with
$Q_k=\{\zc^i{:}y_i{=}k\}$ and $P_k\sim p(\zc\mid y{=}k)$,
$\mathcal{L}_{\mathrm{agg}}=\MMD^2(\{\zc^i\},\{z_{c,r_j}^p\})$ against a
random-class prior mixture, and $\mathcal{L}_{\mathrm{cls}}=\CE(h(\muc),y)$ a
linear auxiliary classifier preventing semantic collapse.

\section{C. Training Hyperparameters and Schedule}

\begin{table}[h]
\centering
\small
\caption{Hyperparameters used in all \ours{} experiments.}
\label{tab:hyperparameters}
\begin{tabular}{ll}
\toprule
Parameter & Value \\
\midrule
Optimizer & Adam, $\mathrm{lr}=10^{-3}$ \\
LR schedule & StepLR, step 100, $\gamma=0.5$ \\
Batch size & 128 \\
Total epochs & 300 \\
Gradient clip (global norm) & 1.0 \\
Semantic dim $d_c$ / style dim $d_s$ & 64 / 128 \\
Prior concentration $\rho_p$ & 0.7 \\
EMA momentum $\tau$ & 0.95 \\
$\lambda_1$ / $\lambda_{\mathrm{LPIPS}}$ & 1.0 / 0.1 \\
$\alpha_f$ / $\beta_f$ (class / agg.\ MMD) & 2.0 / 5.0 \\
$\gamma_f$ (style) & 1.0 \\
$\delta_f$ (per-class style, off/on) & 0 / 1 \\
$\eta_f$ (classifier) & 0.3 \\
Style prior temperature $\tau_s$ & 0.25 \\
\bottomrule
\end{tabular}
\end{table}

\begin{table}[h]
\centering
\scriptsize
\setlength{\tabcolsep}{2pt}
\caption{Training phase boundaries and active loss terms; ramps are linear.}
\label{tab:schedule}
\begin{tabular}{lll}
\toprule
Phase & Epochs & Terms and ramp \\
\midrule
A & 0--49   & $\mathcal{L}_{\mathrm{rec}}$ only \\
B & 50--99  & $+\mathcal{L}_{\mathrm{style}}(0{\to}\gamma_f)$,
              $\mathcal{L}_{\mathrm{style\text{-}cls}}(0{\to}\delta_f)$,
              $\mathcal{L}_{\mathrm{cls}}(0.1{\to}0.2)$ \\
C & 100--199 & $+\mathcal{L}_{\mathrm{class}}(0{\to}\alpha_f/2)$,
               $\mathcal{L}_{\mathrm{agg}}(0{\to}\beta_f/2)$,
               $\mathcal{L}_{\mathrm{cls}}(0.2{\to}0.3)$ \\
D & 200--299 & $\mathcal{L}_{\mathrm{class}}(\alpha_f/2{\to}\alpha_f)$,
               $\mathcal{L}_{\mathrm{agg}}(\beta_f/2{\to}\beta_f)$ \\
\bottomrule
\end{tabular}
\end{table}

Per-class style regularization ramps up in Phase B, before
$\mathcal{L}_{\mathrm{class}}$ and $\mathcal{L}_{\mathrm{agg}}$ are
introduced in Phase C, so that it begins as early as possible relative to the
semantic MMD terms, limiting rather than eliminating the epochs during which
class-specific $(\zc,\zs)$ interactions can form without style-conditional
counter-pressure. Phase A itself (epochs 0--49) is reconstruction-only, so
the decoder does see real $(\zc,\zs)$ pairs before any style or class
regularization is active. The no-warmup perturbation reported in the main
paper tests whether removing this phase changes the leakage and finds that it
does not.

The eight baselines in the main paper's CIFAR-10 competence table---ResNetAE
plus seven label-using methods---use the same ResNet-18 backbone, total latent
dimension ($d{=}192$), 300-epoch budget, and optimizer settings as \ours{}.
The four MNIST cross-model diagnostic controls (VAE, WAE-MMD,
$\beta$-TCVAE, and FactorVAE) instead share a lightweight 3-layer CNN encoder,
latent dimension 64, and a 100-epoch budget. The two groups serve different
comparisons and are not ranked against one another.

\section{D. Diagnostic Protocol}

All leakage diagnostics use $n{=}2048$ held-out test samples with a fixed
seed, so repeated invocations across checkpoints evaluate the same subset
rather than adding sampling noise on top of single-seed training runs. The
Euclidean MMD estimator uses the average of a multi-scale RBF bandwidth
ladder $\sigma\in\{0.5,1,2,5,10,20,50\}$; a single median-heuristic
bandwidth saturates at $\zs$'s empirical scale and returns near-zero
regardless of distributional mismatch. The same saturation is why we do not
report the HSIC proxy: with a single fixed bandwidth it returns a
near-constant value across checkpoints whose $\Dinter$ and LP vary by factors
of five.

JointMMD, the within-class dependence diagnostic defined in the main paper,
uses a product kernel: the multi-scale
spherical RBF above for $\zc$ and the multi-scale Euclidean RBF for $\zs$,
with per-block standardization. The permutation null independently
permutes both sample sets within each class, so any residual value reflects
estimator bias rather than dependence; we report $200$ permutations per
checkpoint across 3 seeds. With the standard plus-one correction, the minimum
attainable value is $1/(200+1)\approx0.005$, so null-exceedance cases are
reported as $p{=}0.005$.

The linear probe is a single linear layer trained on the encoded $\mus$ of
the evaluation split with an 80/20 train/validation split, Adam at $10^{-2}$
for 100 epochs; we report validation accuracy. Because the probe is linear,
its validation accuracy is a conservative proxy for predictability within
the chosen classifier family; it is not presented as a direct numerical
lower bound on mutual information.

The latent-swap intervention draws $n{=}100$ pairs per ordered class pair
($90$ off-diagonal pairs for $K{=}10$), decodes $\mathrm{Dec}(\muc,\mus)$
using posterior means rather than samples to remove reparameterization noise,
and re-encodes the result. We report the diagonal-excluded means of
$P(\mathrm{pred}{=}a)$ (content-following) and $P(\mathrm{pred}{=}b)$
(style-following); these do not sum to one, and the residual is the mass
assigned to the other eight classes.

\section{E. Extended Results}

\subsection{E.1 Full $\delta$ sweep}

Table~\ref{tab:fullsweep} gives the complete six-point sweep including
reconstruction and sample-quality columns omitted from the main paper.
Naive FID improves overall but not monotonically with $\delta$ on MNIST,
from $74.0$ at $\delta{=}0$ to $41.5$ at $\delta{=}3$, with a temporary rise
from $52.6$ to $59.4$ at $\delta{=}0.3$. The
remedy makes naive prior samples \emph{better} even though it does not make
them correct. The two quantities are not the same thing, and conflating them
is precisely the error the paper is about. Reconstruction is essentially
flat across the sweep (SSIM $0.977$--$0.983$ on MNIST), so the remedy is not
purchasing leakage reduction with reconstruction quality.

\begin{table}[h]
\centering
\scriptsize
\setlength{\tabcolsep}{3pt}
\caption{Full $\delta$ sweep, single seed per point (measured).}
\label{tab:fullsweep}
\begin{tabular}{llrrrrr}
\toprule
Data & $\delta$ & ACC & $\Dinter$ & LP & FID & SSIM \\
\midrule
MNIST & 0    & .9915 & 6.21 & 99.9\% & 74.0 & .983 \\
MNIST & 0.03 & .9944 & 3.27 & 99.8\% & 62.8 & .983 \\
MNIST & 0.1  & .9898 & 2.12 & 99.3\% & 52.6 & .983 \\
MNIST & 0.3  & .9936 & 1.55 & 84.5\% & 59.4 & .982 \\
MNIST & 1    & .8592 & 1.22 & 42.6\% & 43.4 & .979 \\
MNIST & 3    & .9929 & 1.18 & 43.0\% & 41.5 & .977 \\
\midrule
CIFAR & 0    & .8129 & 3.74 & 81.7\% & 80.5 & .706 \\
CIFAR & 0.03 & .8111 & 2.48 & 69.3\% & 79.1 & .707 \\
CIFAR & 0.1  & .8100 & 2.03 & 58.5\% & 77.5 & .710 \\
CIFAR & 0.3  & .7148 & 1.59 & 51.3\% & 72.0 & .707 \\
CIFAR & 1    & .8000 & 1.05 & 30.2\% & 74.4 & .708 \\
CIFAR & 3    & .7994 & 0.95 & 28.5\% & 76.1 & .702 \\
\bottomrule
\end{tabular}
\end{table}

Note that the $\Dinter$ values here are computed on the $\delta$-sweep
checkpoints and differ in the third significant figure from the values in
the main paper's cross-model diagnostic table, which come from separately
trained
$\delta{=}0$/$\delta{=}1$ checkpoints used for the cross-model comparison
($6.21$ vs $6.27$ at $\delta{=}0$; both round to $1.22$ at $\delta{=}1$).
This is ordinary run-to-run variation between independently trained models
at the same setting and illustrates why the single-seed caveat matters.

\begin{figure}[t]
\centering
\includegraphics[width=\columnwidth]{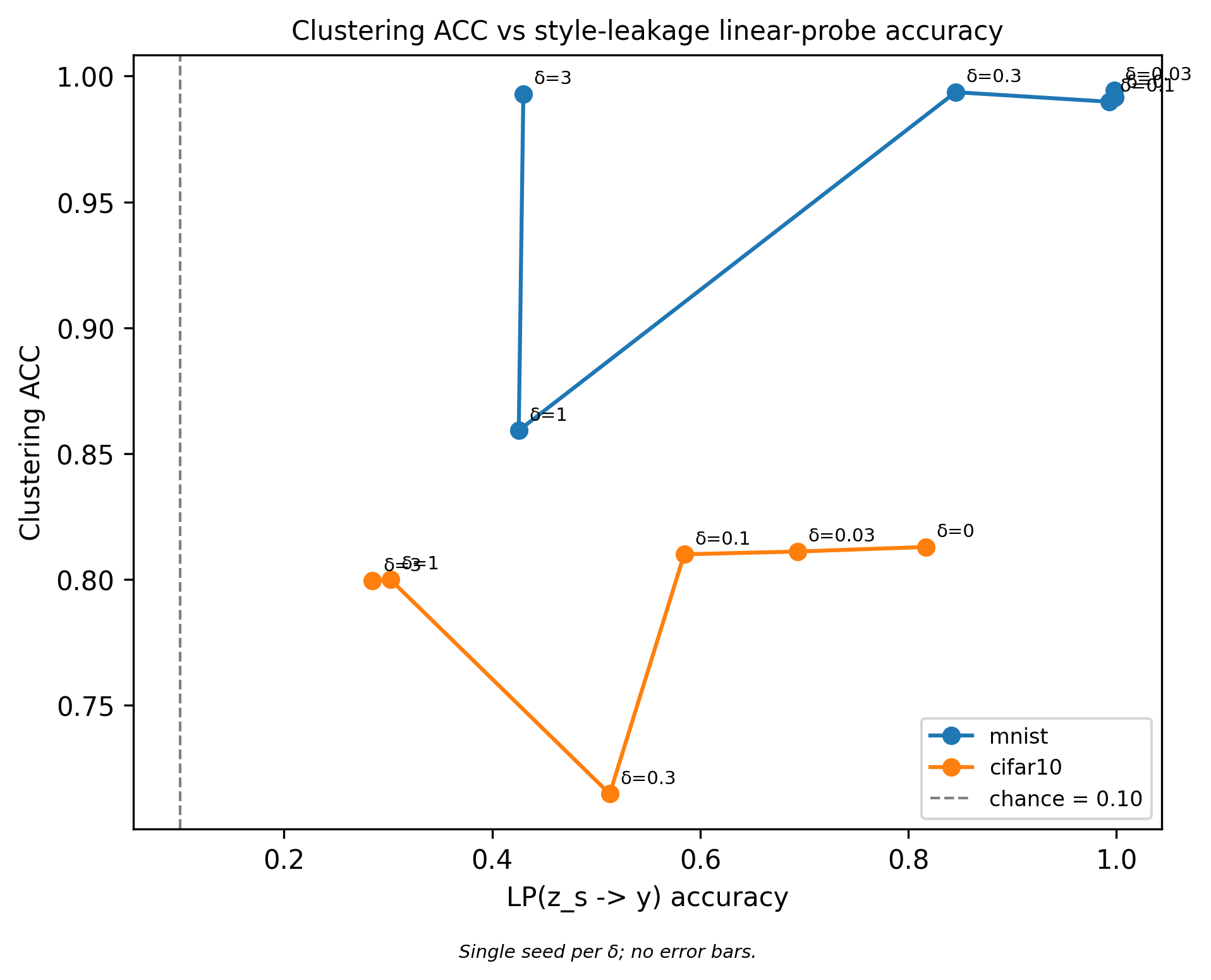}
\caption{Clustering ACC against linear-probe leakage across the six-point
$\delta$ sweep, both datasets (single seed per point; the connecting line is
a visual aid, not a claim of a smooth underlying curve). Leakage (the
$x$-axis) moves smoothly with $\delta$; ACC does not.}
\label{fig:pareto-acc}
\end{figure}

\begin{figure}[t]
\centering
\includegraphics[width=\columnwidth]{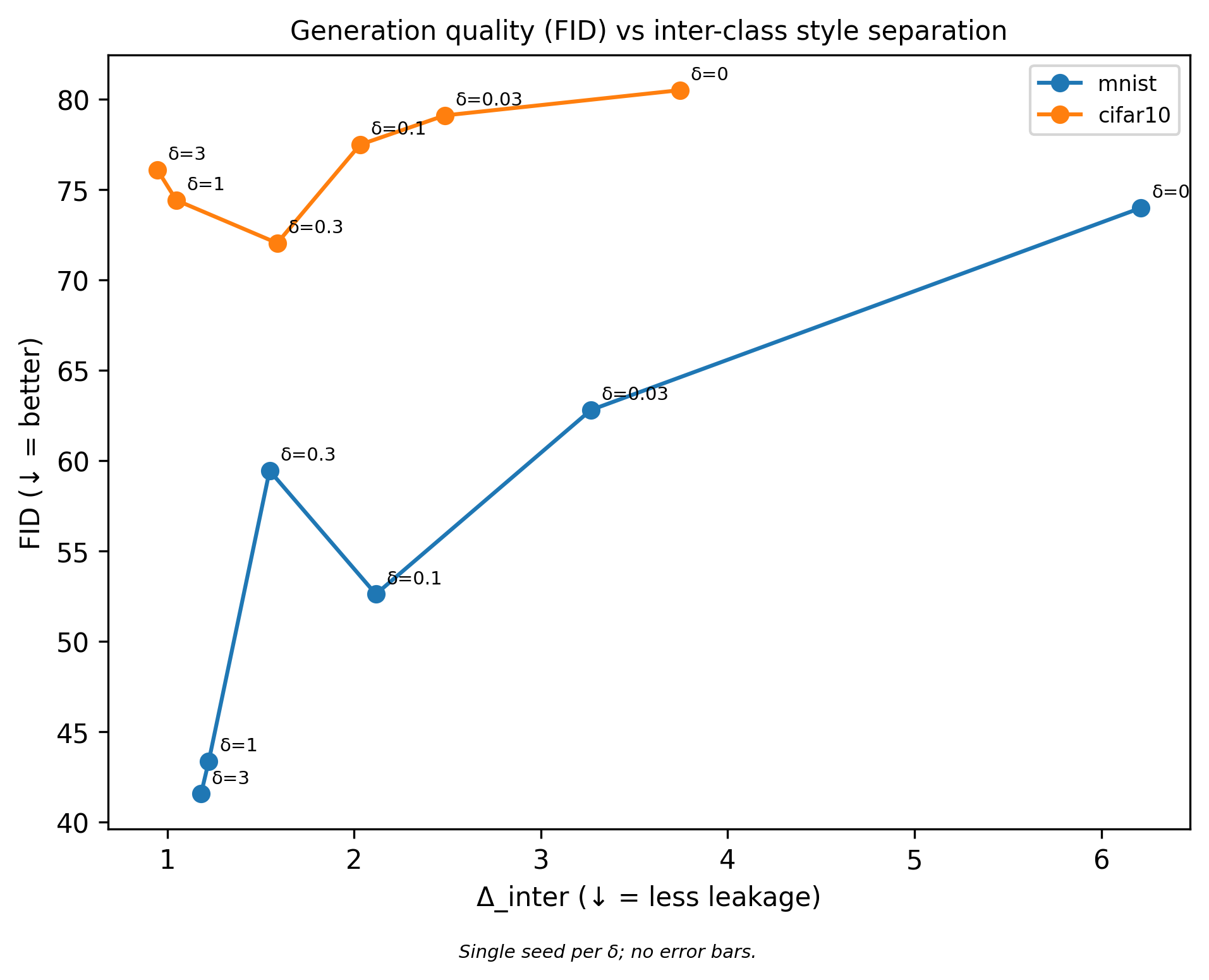}
\caption{Naive-sampling FID against $\Dinter$ across the same sweep.
$\Dinter$ falls steadily, whereas FID improves overall but non-monotonically;
lower leakage tends to accompany better naive sample quality without making
sampling \emph{correct}.}
\label{fig:pareto-fid}
\end{figure}

\subsection{E.2 Cross-dataset leakage}

\begin{table}[h]
\centering
\small
\caption{Leakage across three datasets at $\delta{=}0$ (measured),
ordered by measured intra-class diversity.}
\label{tab:crossdataset-app}
\begin{tabular}{lrrr}
\toprule
Dataset & Diversity & $\Dinter$ & LP \\
\midrule
MNIST         & 0.33 & 6.27 & 100.0\% \\
Fashion-MNIST & 0.47 & 4.86 & 97.7\% \\
CIFAR-10      & 0.66 & 3.80 & 86.9\% \\
\bottomrule
\end{tabular}
\end{table}

\begin{figure}[t]
\centering
\includegraphics[width=\columnwidth]{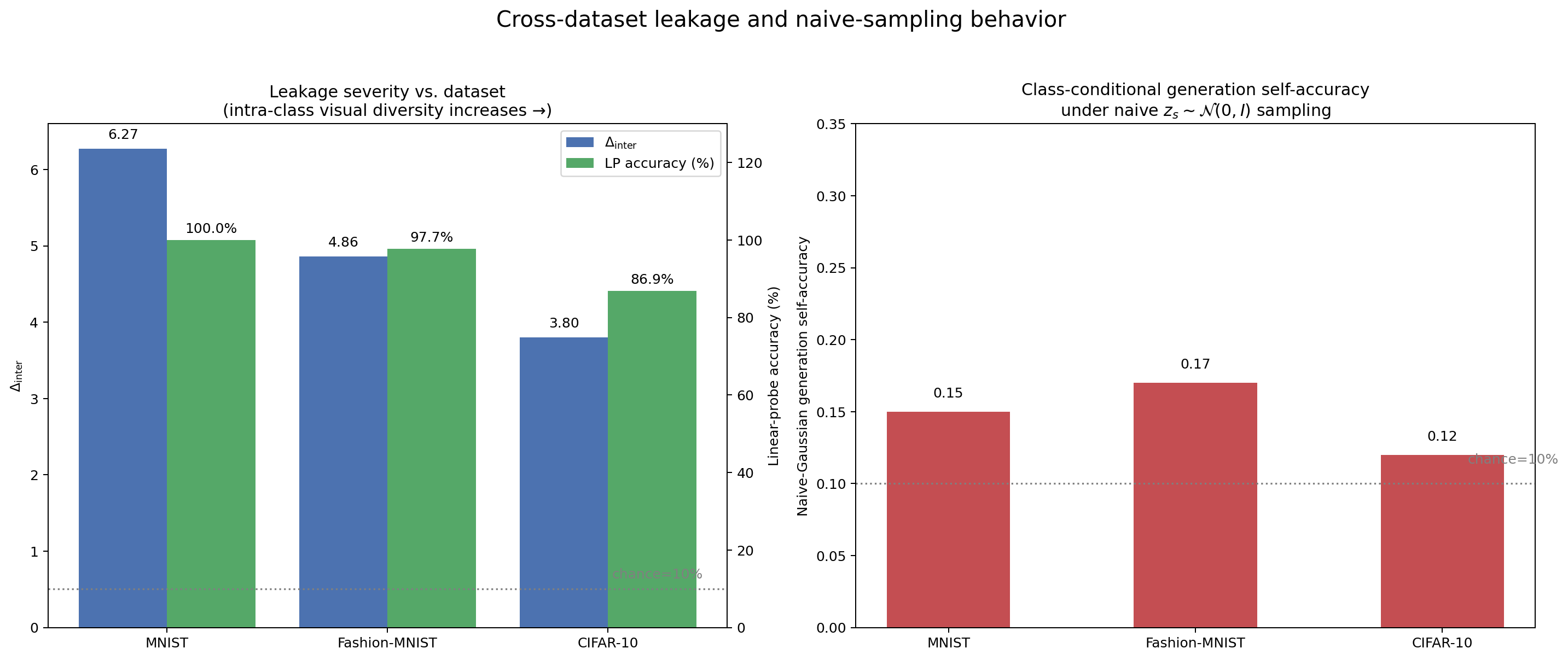}
\caption{Left: $\Dinter$ and linear-probe accuracy across the three
datasets, ordered by intra-class visual diversity; both decrease
monotonically. Right: naive-Gaussian generation self-accuracy does
\emph{not} follow the same order (CIFAR-10 is lowest, not highest), because
it additionally inherits dataset-intrinsic classification difficulty; see
the main paper's discussion.}
\label{fig:crossdataset}
\end{figure}

Fashion-MNIST at $\delta{=}0$ reaches ACC $0.9341$, NMI $0.8672$, ARI
$0.8633$, and FID $72.37$; its leakage metrics lie between MNIST and
CIFAR-10, while its FID is slightly lower than MNIST's
(Figure~\ref{fig:crossdataset}). On
MNIST the single-latent CS-WAE variant (one Spherical Cauchy semantic latent,
no style variable) reaches ACC $0.8735$, NMI $0.8863$, ARI $0.8411$, FID
$26.95$. It is worth noting explicitly that this ablation has \emph{no style
latent and therefore cannot leak by construction}, and it also achieves the
best FID of any model we trained. That is not an argument for removing the
style variable, since the factorization exists to provide independent style
control, but it does bound how much the factorization is buying on this
dataset.

\subsection{E.3 Per-class conditional MMD}

\begin{figure}[t]
\centering
\includegraphics[width=0.49\columnwidth]{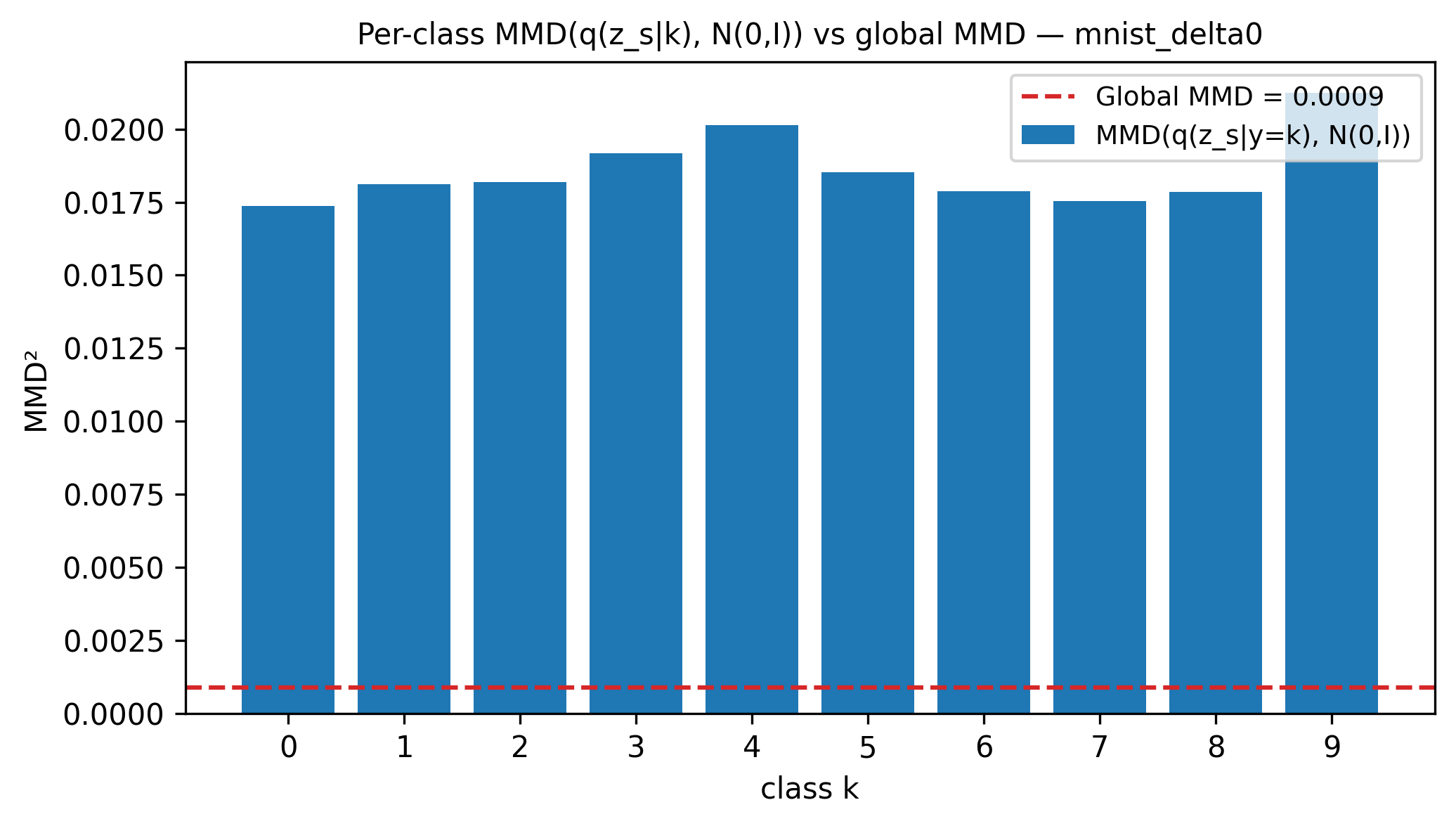}\hfill
\includegraphics[width=0.49\columnwidth]{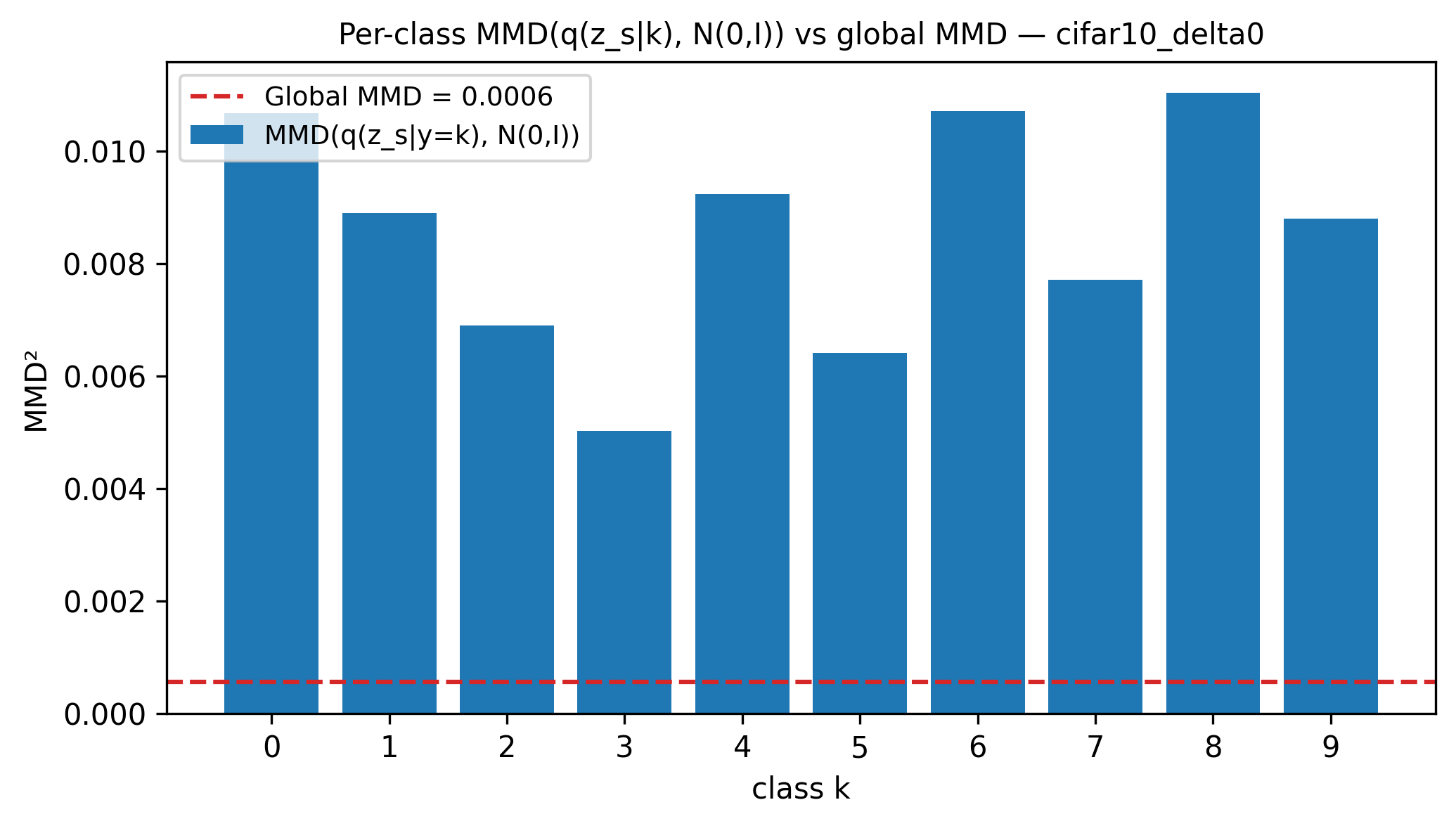}
\caption{Per-class $\MMD^2(q(\zs\mid y{=}k),\mathcal N(0,I))$ at
$\delta{=}0$ (bars) against the global MMD the marginal regularizer actually
optimizes (dashed line). Left: MNIST; right: CIFAR-10. Every class sits far
above the line the model was trained to minimize.}
\label{fig:cmmd-bars}
\end{figure}

\begin{figure}[t]
\centering
\includegraphics[width=0.49\columnwidth]{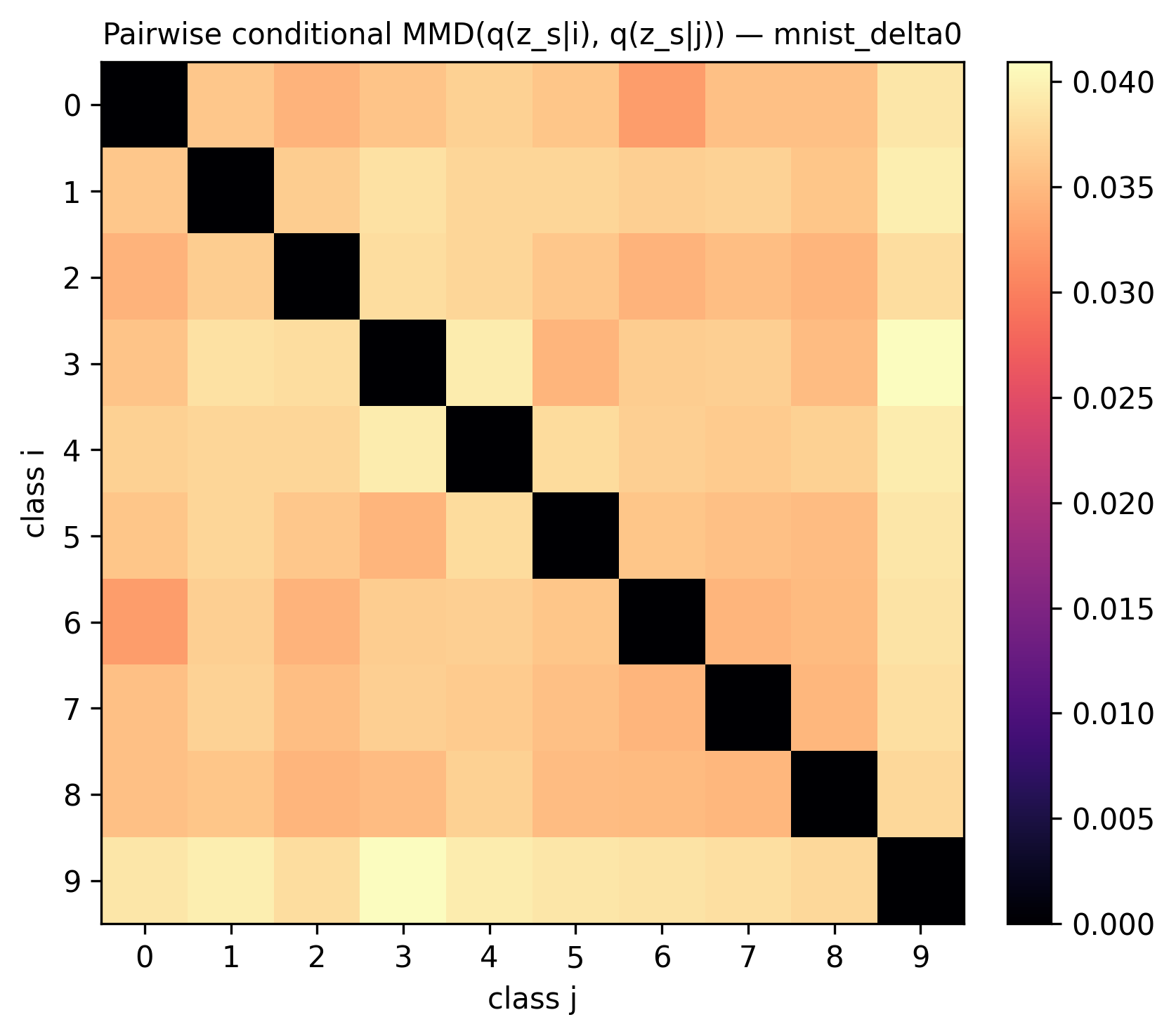}\hfill
\includegraphics[width=0.49\columnwidth]{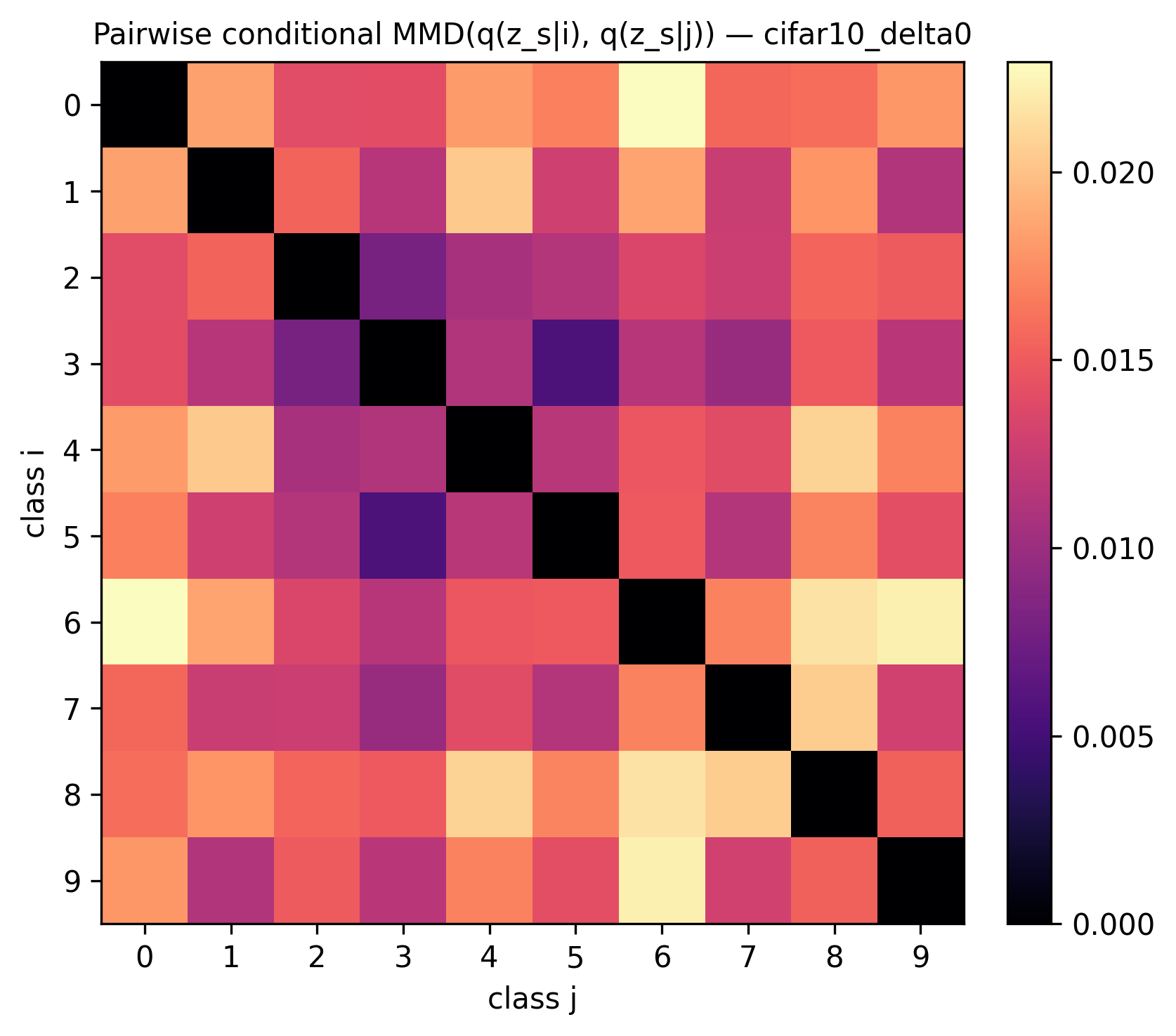}
\caption{Pairwise $\MMD^2(q(\zs\mid y{=}i),q(\zs\mid y{=}j))$ between all
class pairs at $\delta{=}0$ (left MNIST, right CIFAR-10). Broad off-diagonal
structure, not a few isolated confusable pairs.}
\label{fig:cmmd-pairwise}
\end{figure}

The single scalar $\Dinter$ compresses a $K$-way structure. Computing
$\MMD^2(q(\zs\mid y{=}k),\mathcal N(0,I))$ per class at $\delta{=}0$ gives a
mean of $0.0186$ on MNIST against a global MMD of $0.00089$, a factor of
$21$; on CIFAR-10, $0.0085$ against $0.00056$, a factor of $15$. Every one of
the ten per-class values on both datasets sits above the global figure, so
this is not a few confusable classes dragging an average: every class
contributes (Figure~\ref{fig:cmmd-bars}). At $\delta{=}1$ the ratios fall to
$4.4\times$ (MNIST) and $5.3\times$ (CIFAR-10) without reaching one. The
pairwise structure $\MMD^2(q(\zs\mid y{=}i),q(\zs\mid y{=}j))$ likewise shows
broad off-diagonal separation on both datasets rather than a few isolated
pairs (Figure~\ref{fig:cmmd-pairwise}).

\begin{figure}[t]
\centering
\includegraphics[width=\columnwidth]{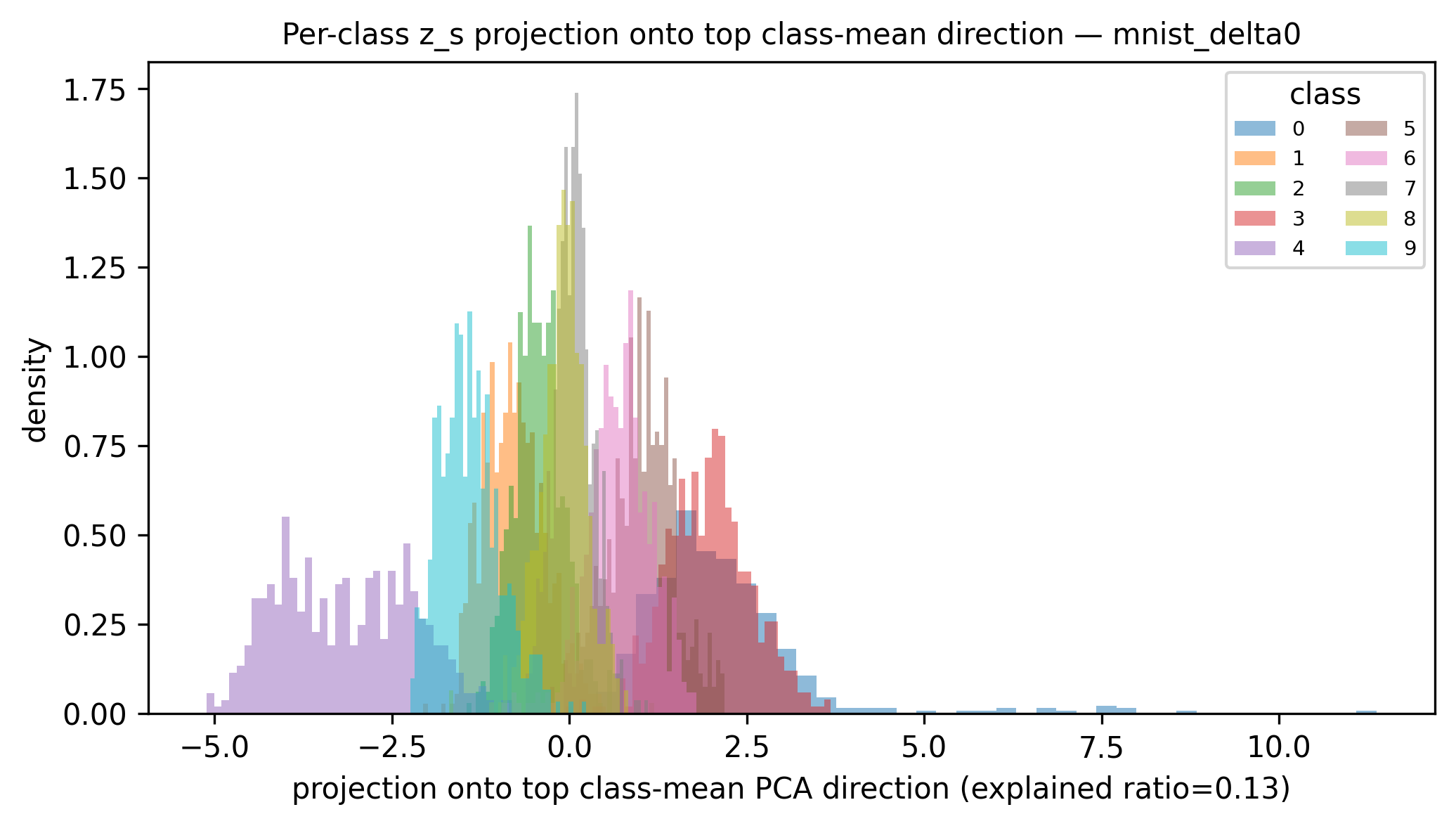}
\caption{Per-class histograms of $\zs$ projected onto the top principal
component of the class-mean matrix (MNIST, $\delta{=}0$). Some classes
separate cleanly along this single direction even where a 2D t-SNE does not
make the separation visible, illustrating that a fixed low-dimensional
projection can either overstate or understate leakage.}
\label{fig:cmmd-hist}
\end{figure}

\section{F. Failure-Mode Analysis}

\begin{figure}[t]
\centering
\includegraphics[width=\columnwidth]{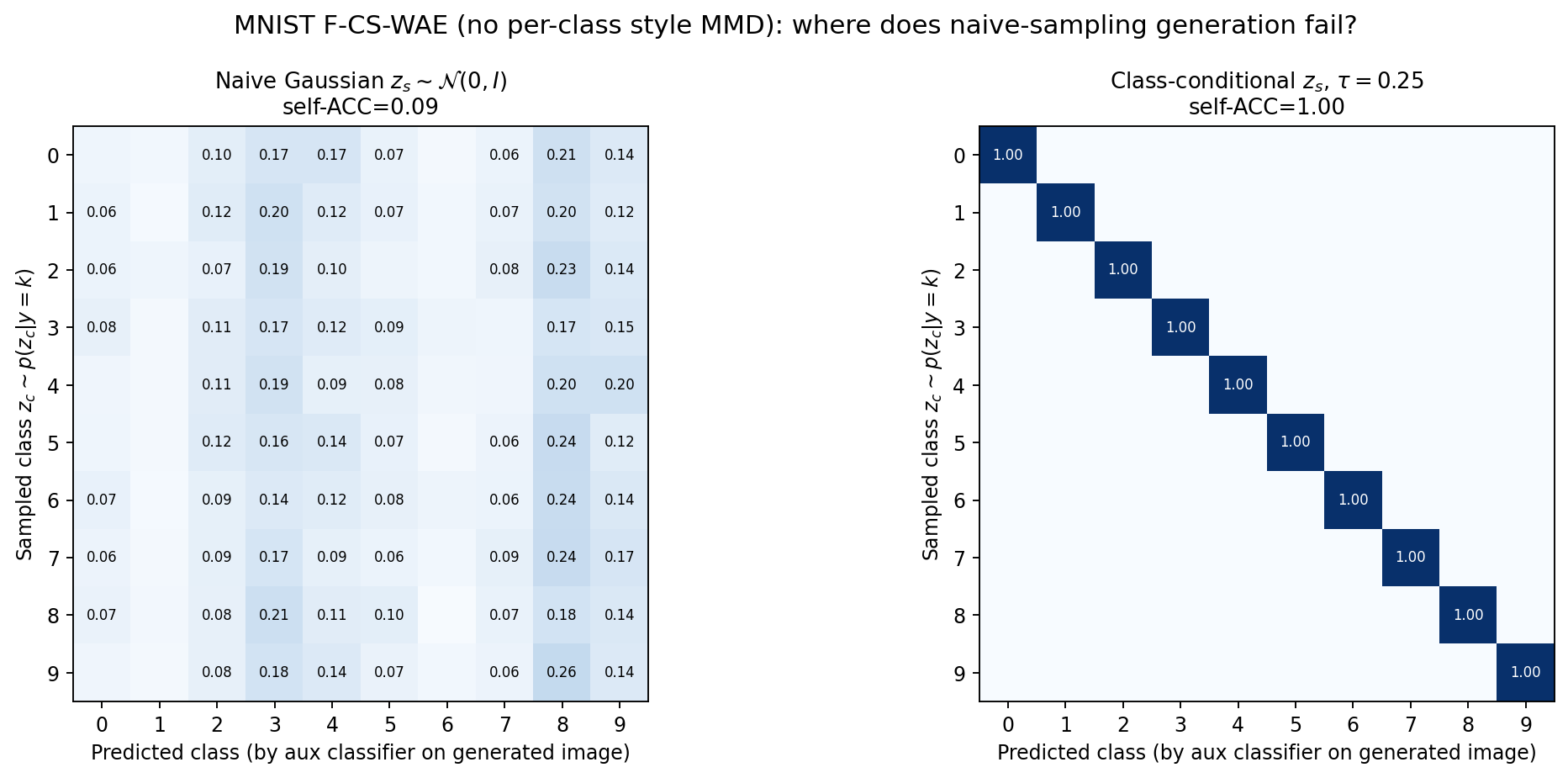}
\caption{Model-internal robustness check: confusion between the sampled
class and the class assigned after re-encoding the generated image. Naive
Gaussian sampling (left) collapses onto attractor digits (notably 8), while
the class-conditional prior (right, $\tau{=}0.25$) is nearly diagonal. The
main paper uses an independent external classifier for primary Gen-ACC.}
\label{fig:confusion}
\end{figure}

The naive-sampling failure is systematic rather than random. Confusion
between the sampled class $y{=}k$ and the class the classifier assigns to the
generated image (Figure~\ref{fig:confusion}) shows probability mass
collapsing onto a small set of attractor digits, notably $8$, largely
independent of which class was requested. This is what one expects if the
decoder resolves an out-of-training-distribution $(\zc,\zs)$ pair by falling
back on whichever class its style input most resembles, and it explains why
model-internal self-accuracy sits near $0.15$ rather than near the $0.10$
that uniform random errors would produce: the errors are concentrated, not
spread. The main paper reports the corresponding external Gen-ACC.

\begin{figure}[t]
\centering
\includegraphics[width=0.49\columnwidth]{figures/mnist_swap_grid_delta0.png}\hfill
\includegraphics[width=0.49\columnwidth]{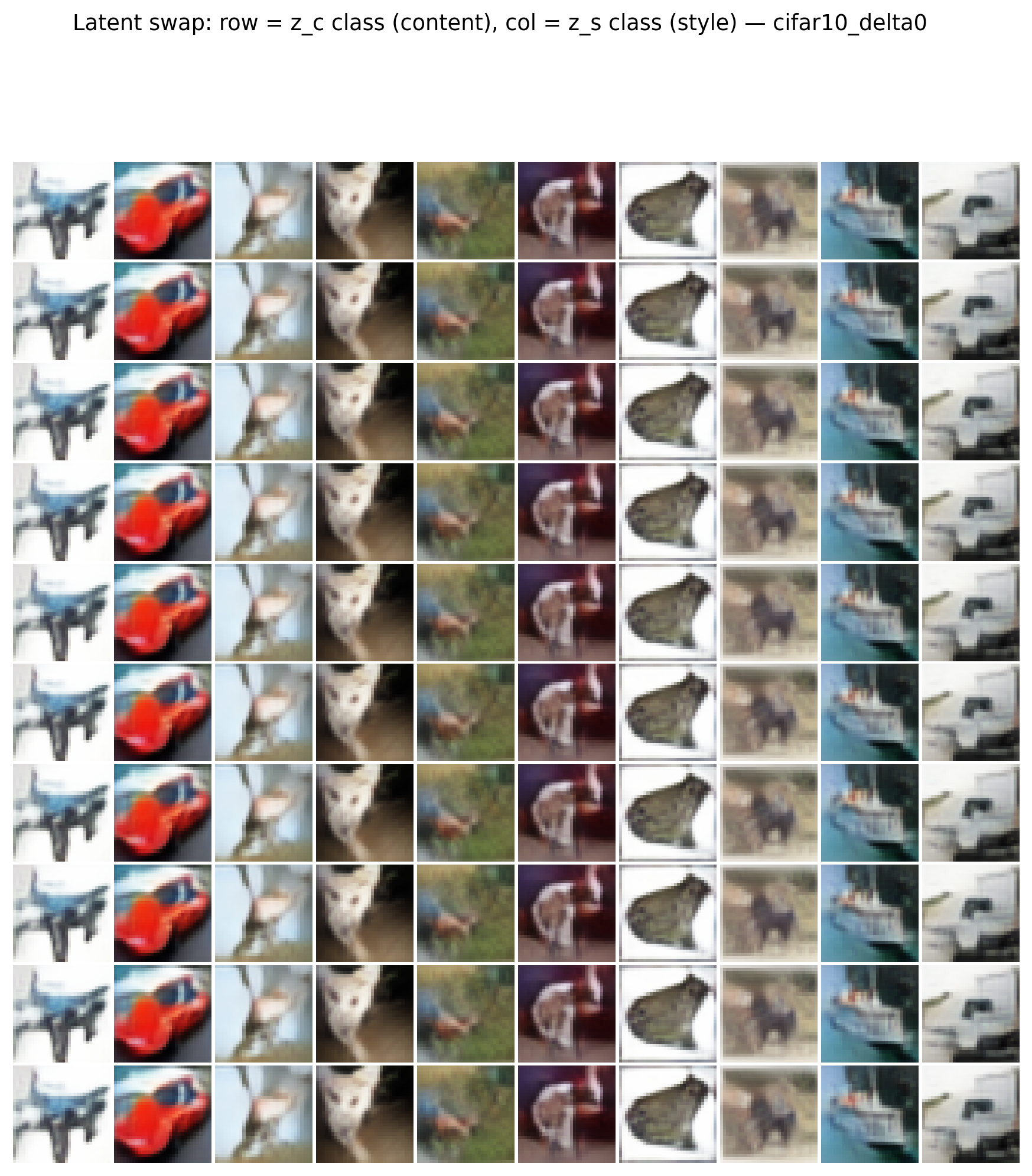}
\caption{Latent-swap grids at $\delta{=}0$: row $=\zc$ donor class, column
$=\zs$ donor class. Left, MNIST: every row is visually identical; identity
is set entirely by the column. Right, CIFAR-10: weaker but still visible
column-driven drift.}
\label{fig:swapgrids}
\end{figure}

\begin{figure}[t]
\centering
\includegraphics[width=0.49\columnwidth]{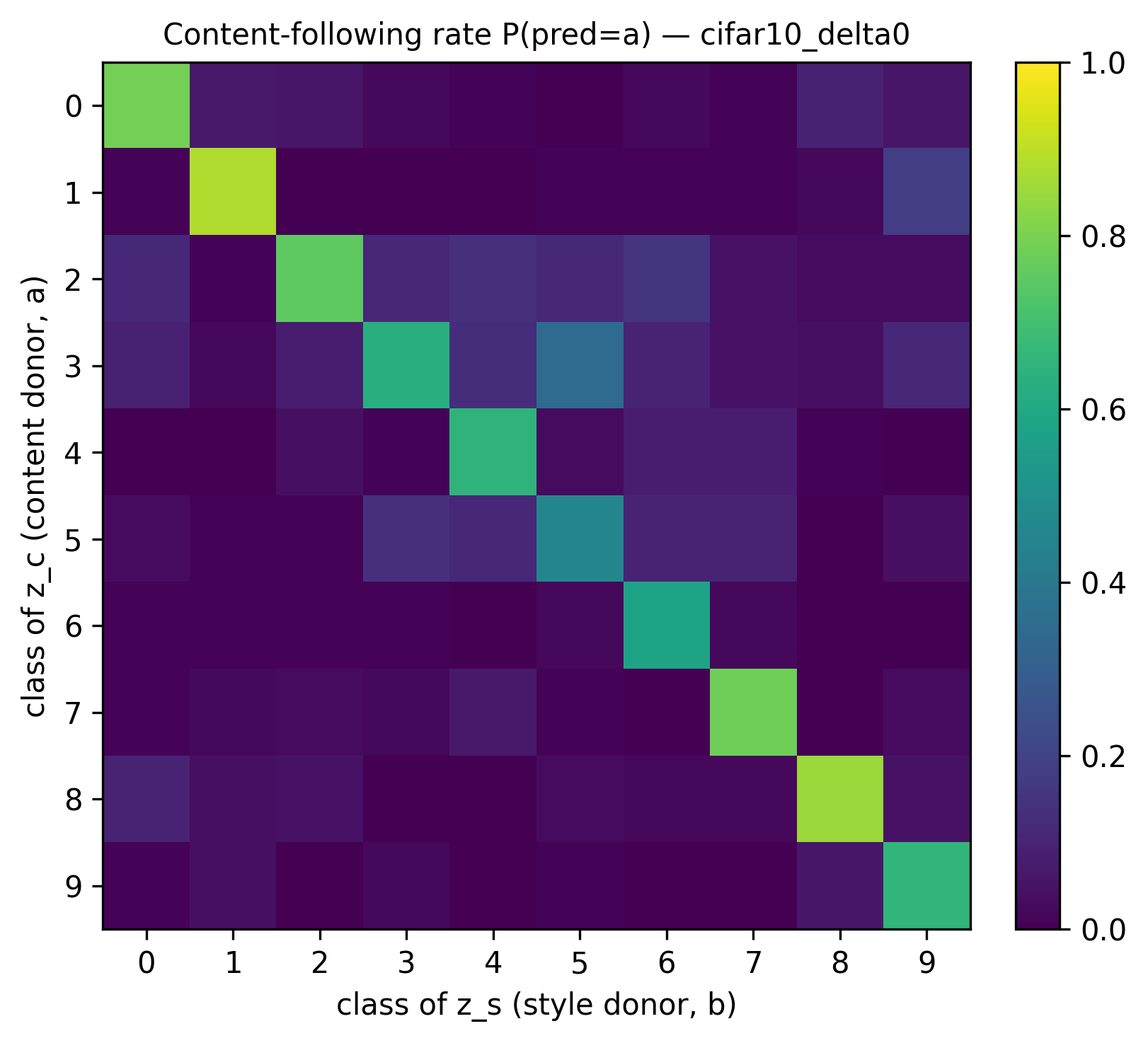}\hfill
\includegraphics[width=0.49\columnwidth]{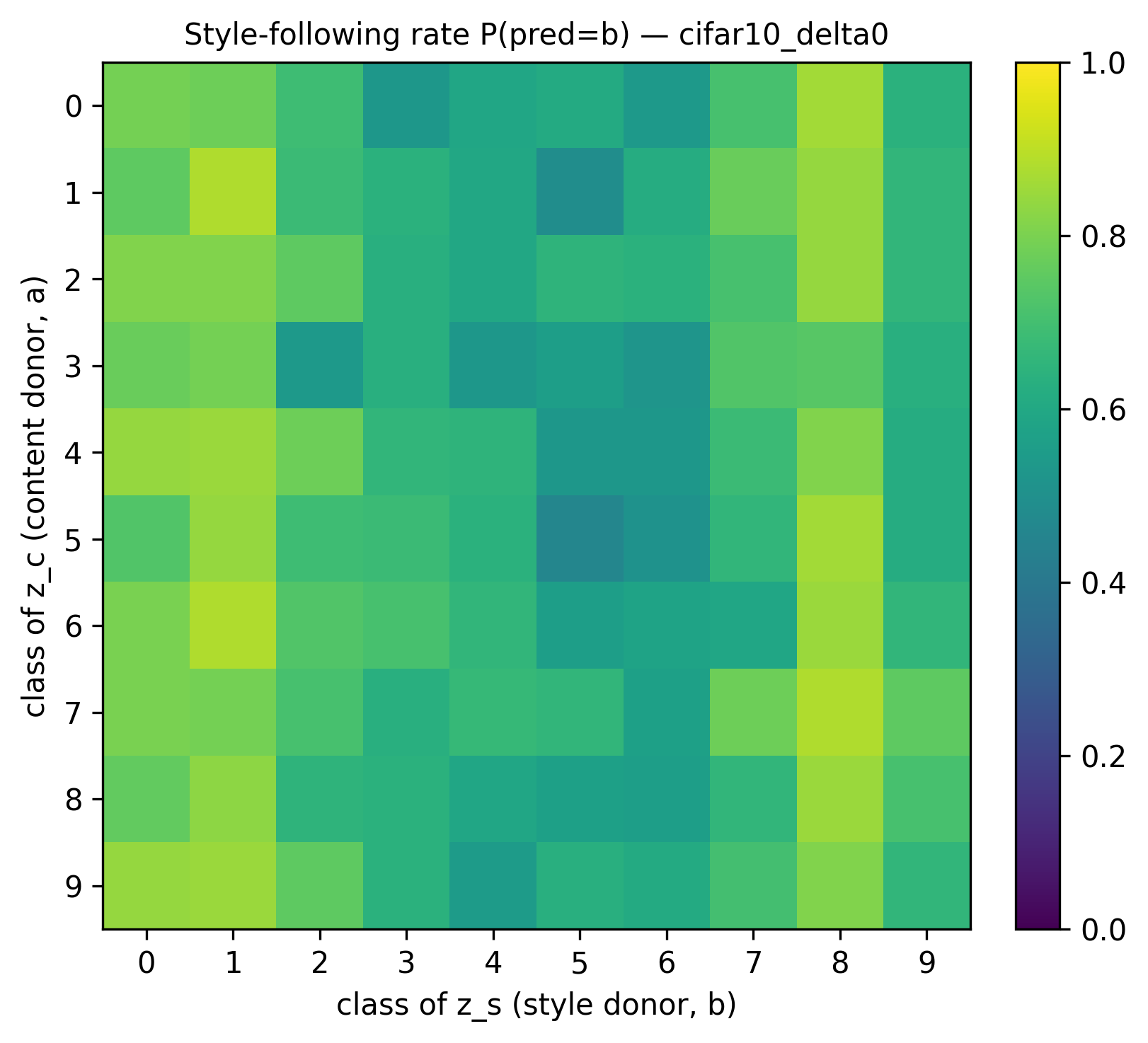}\\
\includegraphics[width=0.49\columnwidth]{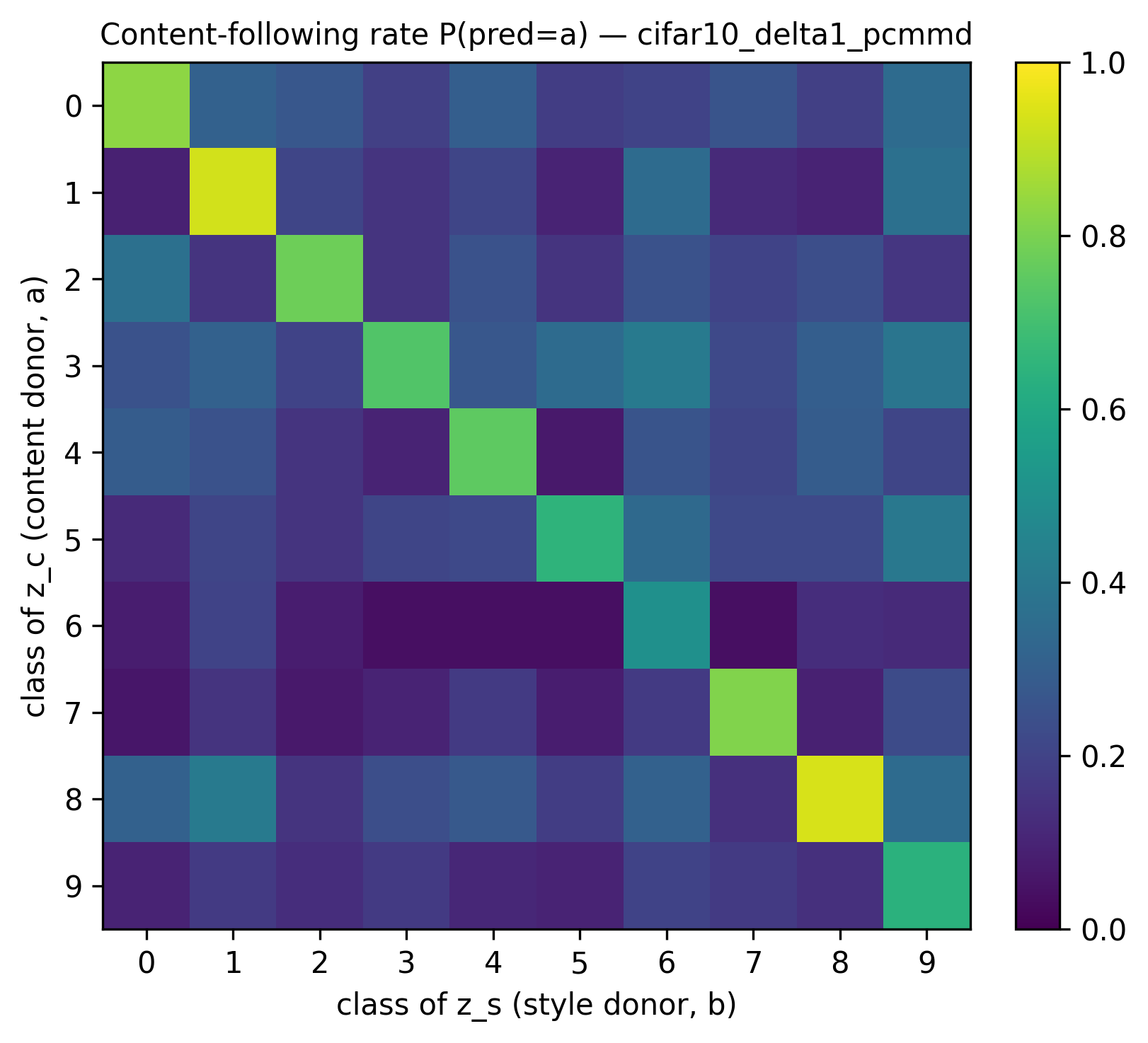}\hfill
\includegraphics[width=0.49\columnwidth]{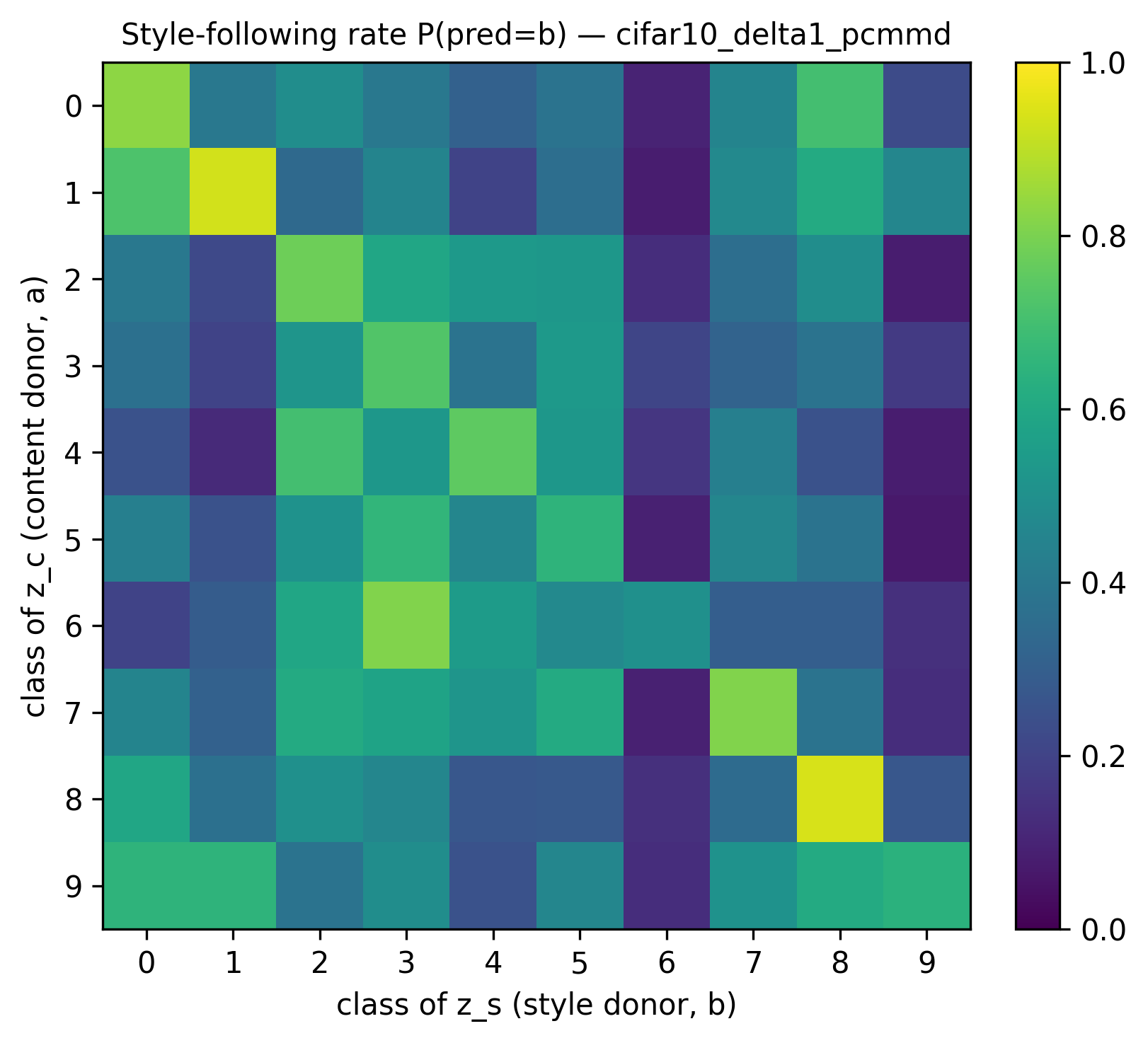}
\caption{Model-internal robustness check: CIFAR-10 latent-swap heatmaps
(left content-following, right style-following; top $\delta{=}0$, bottom
$\delta{=}1$). Primary swap rates in the main paper use the external
classifier.}
\label{fig:cifar-swapheat}
\end{figure}

The latent-swap grid makes the same point at the pixel level
(Figure~\ref{fig:swapgrids}). At $\delta{=}0$ every row (fixed $\zc$ donor)
produces visually identical output; only the column ($\zs$ donor) determines
the digit. On CIFAR-10 the rows are more similar to each other than in the
MNIST grid but still show column-driven drift in object identity for several
classes, consistent with a decoder that was never as exclusively dependent on
$\zs$. Under the model-internal robustness evaluator, the CIFAR-10 swap
heatmaps (Figure~\ref{fig:cifar-swapheat}) show the same
$\delta{=}0\to\delta{=}1$ flip as MNIST but weaker in magnitude
(content-following $4.2\%\to20.0\%$, style-following $68.8\%\to38.5\%$).

\begin{figure*}[t]
\centering
\includegraphics[width=0.98\textwidth]{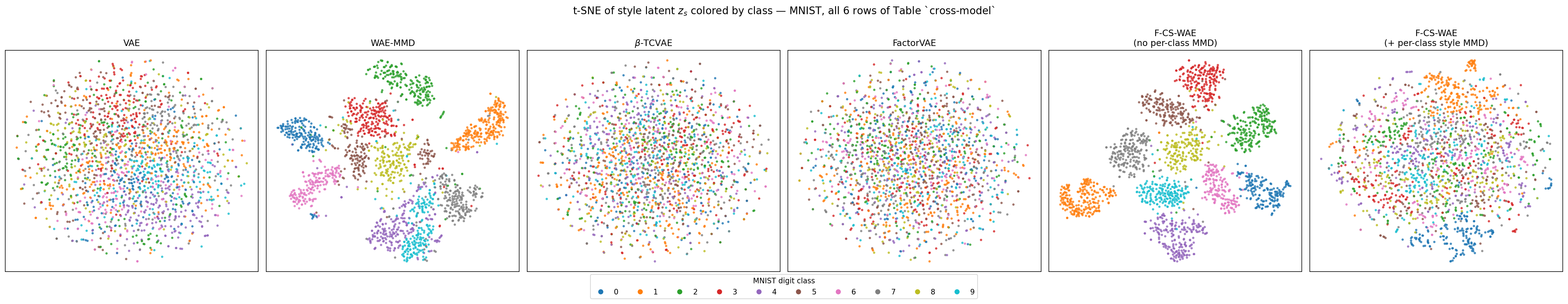}
\caption{t-SNE of the style latent $\zs$, colored by MNIST digit class, for
all six rows of the cross-model table. Color-separated clusters here mean
\emph{worse} (class information leaking into $\zs$): separation is visually
obvious only for WAE-MMD and \ours{} without the remedy, and visibly reduced
after adding per-class style MMD; for the other three families the leakage
is statistically real but invisible in this projection.}
\label{fig:tsne}
\end{figure*}

A 2D t-SNE of $\zs$ colored by class, across all six rows of the cross-model
table (Figure~\ref{fig:tsne}), is worth reading with a caution attached:
unlike a typical clustering figure, \emph{color-separated clusters here mean
worse}. Separation is visually obvious only for WAE-MMD and \ours{} without
the remedy, the two largest-$\Dinter$ rows, and visibly reduced after adding
per-class style MMD. For VAE, $\beta$-TCVAE, and FactorVAE the leakage is
statistically real (LP $74$--$87\%$) but not visually apparent in the
projection. Projecting $\zs$ onto the top principal component of the
$K\times d_s$ class-mean matrix, the same signal $\Dinter$ summarizes,
separates some classes cleanly even where t-SNE does not
(Figure~\ref{fig:cmmd-hist}). A fixed low-dimensional projection can
therefore both overstate and understate leakage depending on which
directions it happens to preserve, which is the argument for the
quantitative diagnostics over any single plot.

\section{G. Sampling Strategies: Qualitative Results}

\begin{figure}[t]
\centering
\includegraphics[width=\columnwidth]{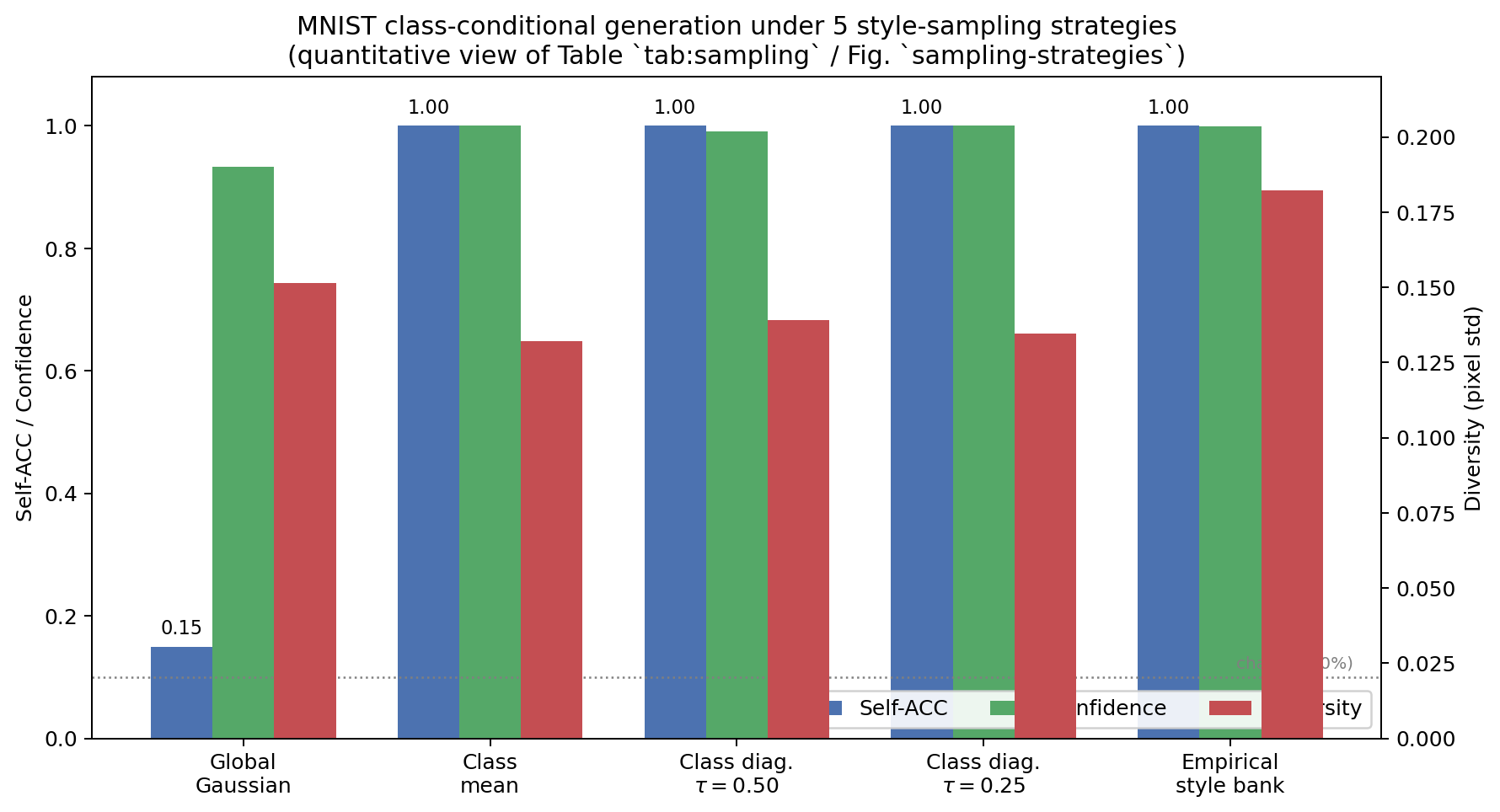}
\caption{Model-internal robustness check for MNIST sampling strategies.
Internal self-accuracy jumps from near-chance to $100\%$ when $\zs$ is
conditioned on class, while diversity is lowest for the class-mean point and
highest for the empirical style bank. External Gen-ACC is reported in the
main paper.}
\label{fig:mnist-bars}
\end{figure}

\begin{figure}[t]
\centering
\includegraphics[width=0.24\columnwidth]{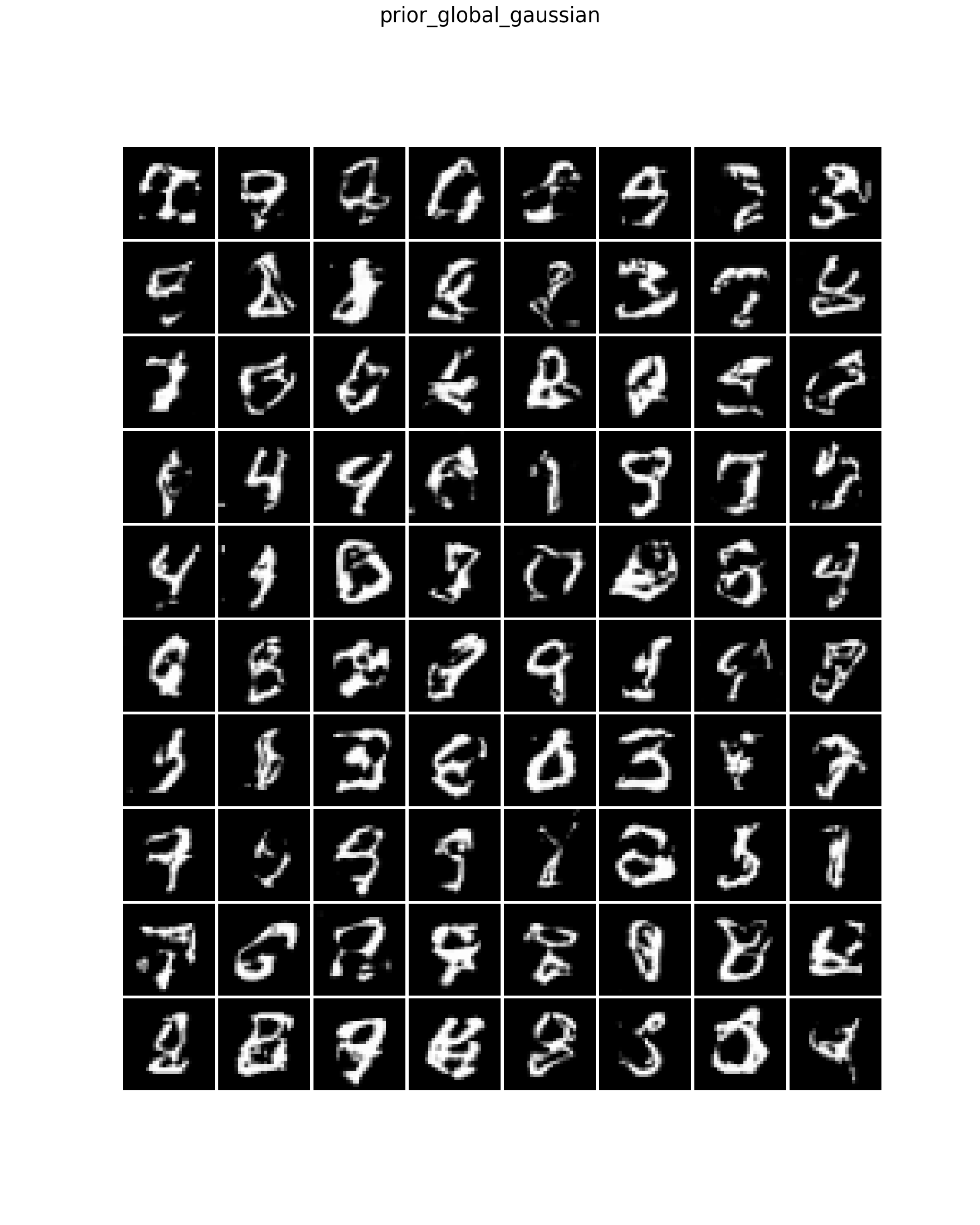}\hfill
\includegraphics[width=0.24\columnwidth]{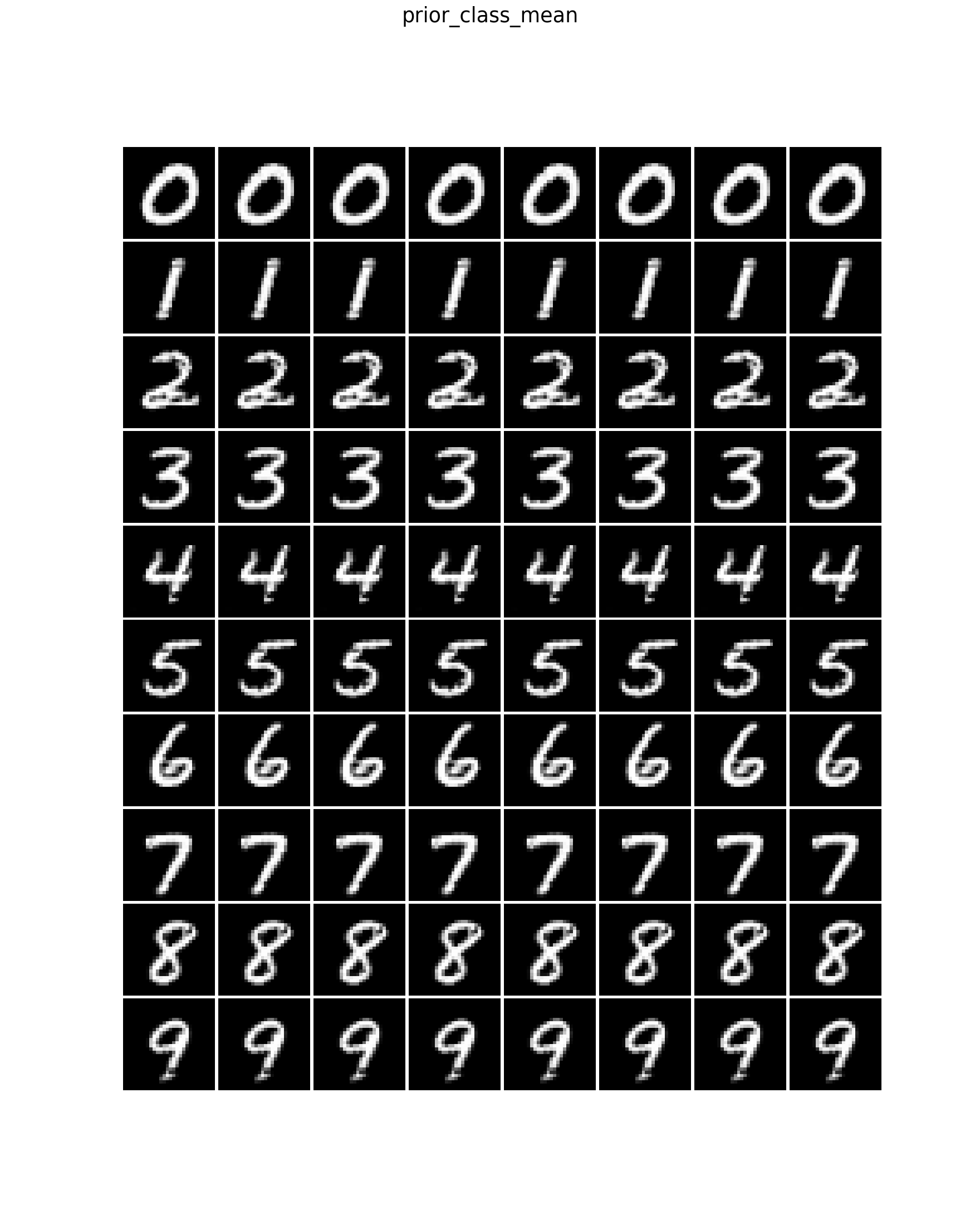}\hfill
\includegraphics[width=0.24\columnwidth]{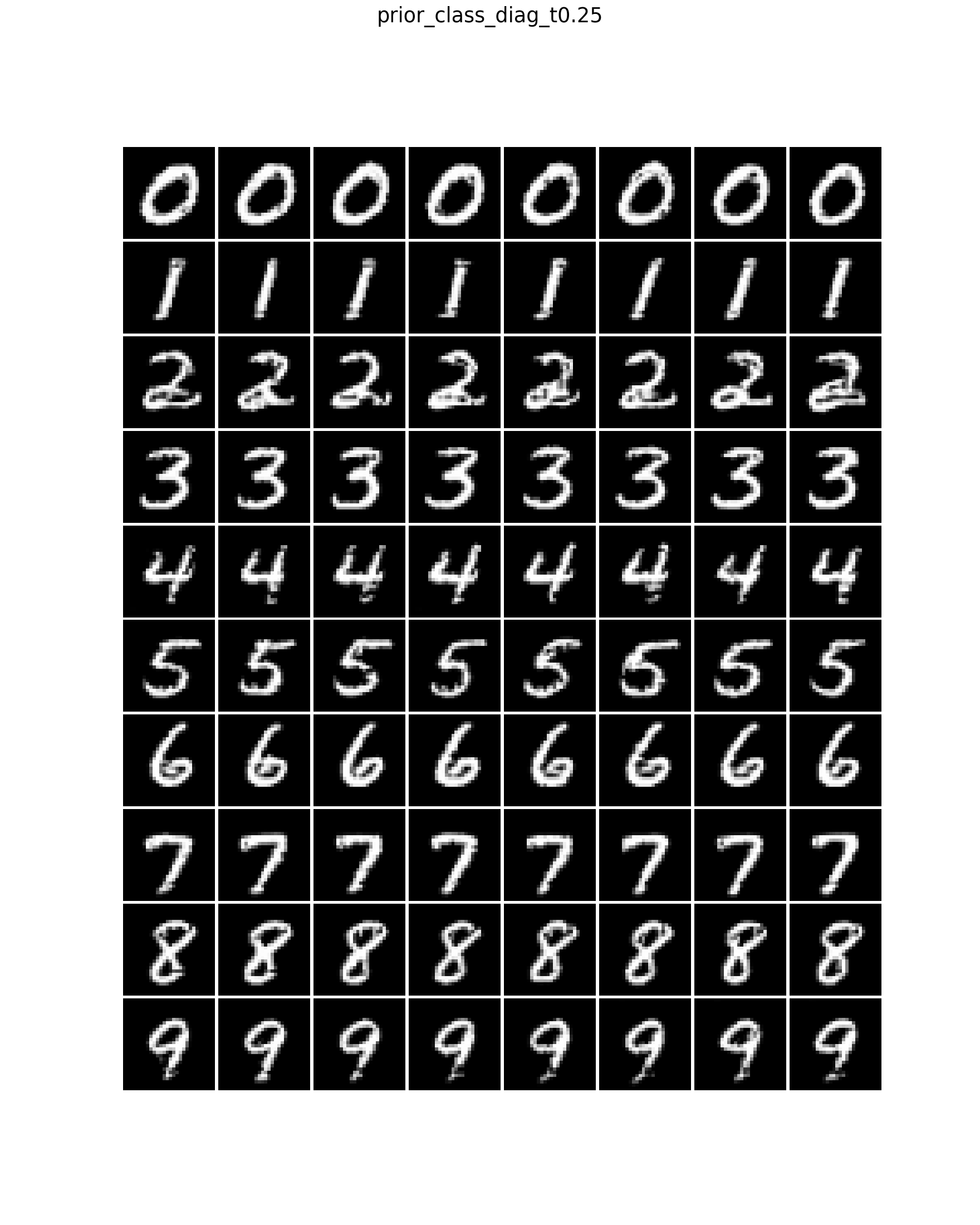}\hfill
\includegraphics[width=0.24\columnwidth]{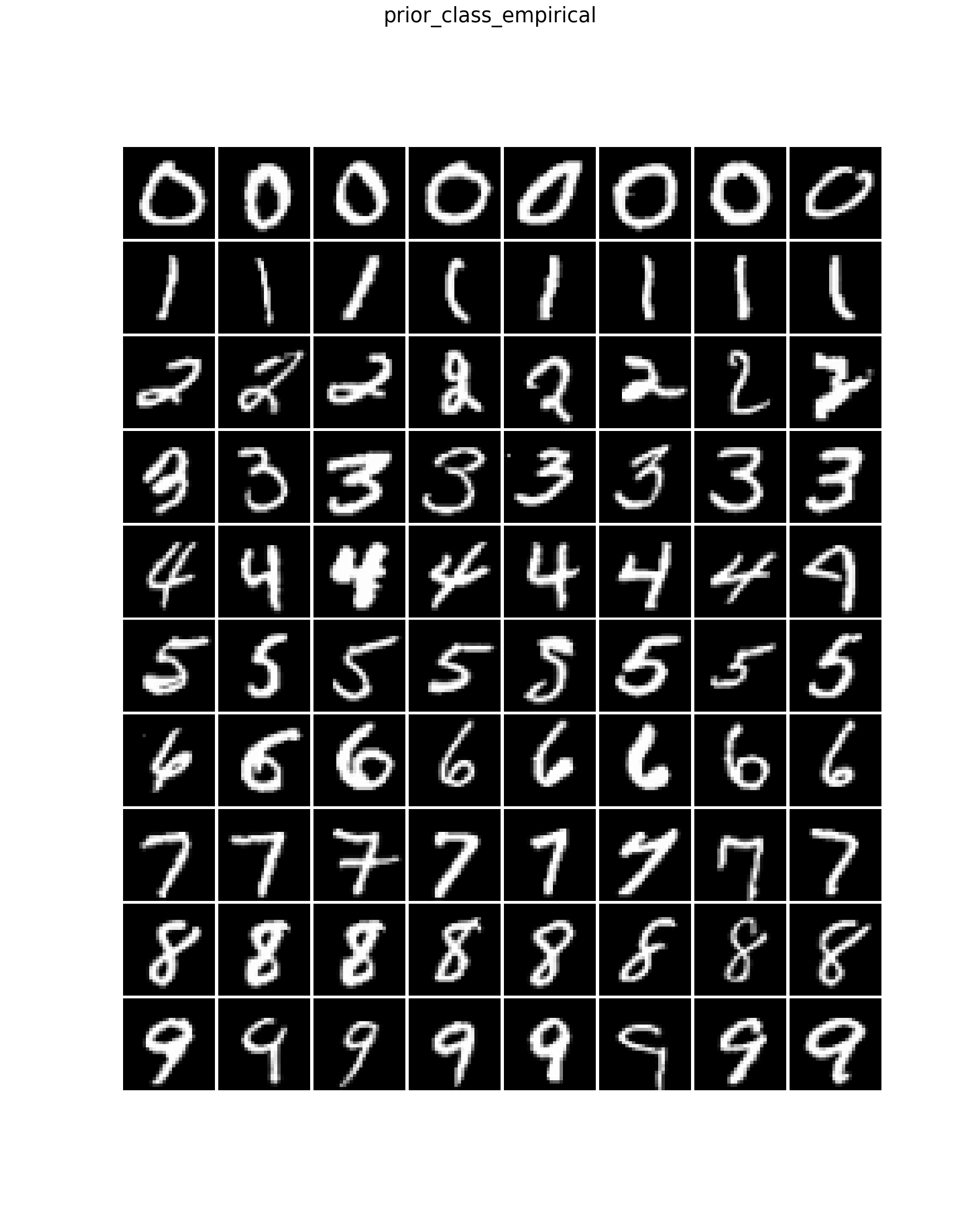}
\caption{MNIST generation under four style-sampling strategies (left to
right): global Gaussian, class mean, class-conditional diagonal
$\tau{=}0.25$, and empirical bank. Under the model-internal robustness
evaluator their self-accuracies are $15\%$, $100\%$, $100\%$, and $100\%$;
these are not the external Gen-ACC values used in the main paper.}
\label{fig:mnist-samples}
\end{figure}

\begin{figure}[t]
\centering
\includegraphics[width=\columnwidth,height=0.72\textheight,keepaspectratio]{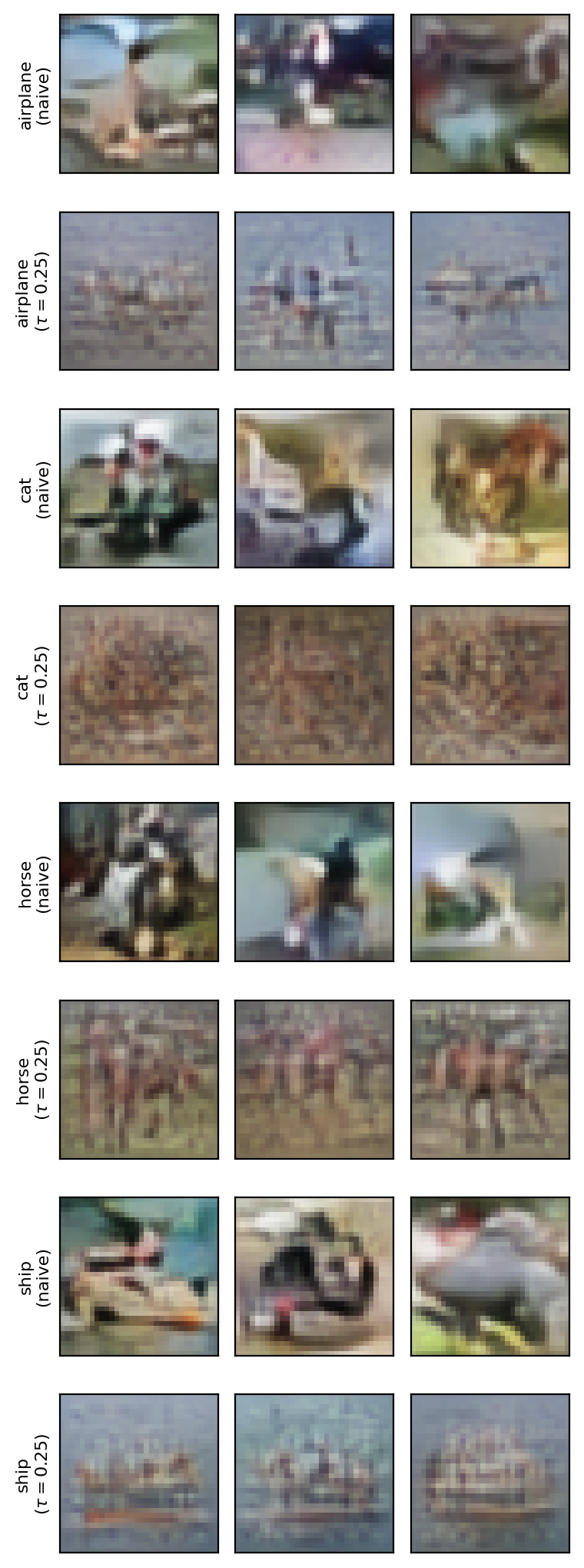}
\caption{CIFAR-10 naive (odd rows) vs.\ class-conditional $\tau{=}0.25$
(even rows) samples, four classes. Diversity within a row collapses under
$\tau{=}0.25$, and identity is often unclear by eye even where the
model-internal evaluator scores it correctly ($0.681$). The corresponding
primary external Gen-ACC in the main paper is $0.41$.}
\label{fig:cifar-grid}
\end{figure}

\begin{figure}[t]
\centering
\includegraphics[width=\columnwidth]{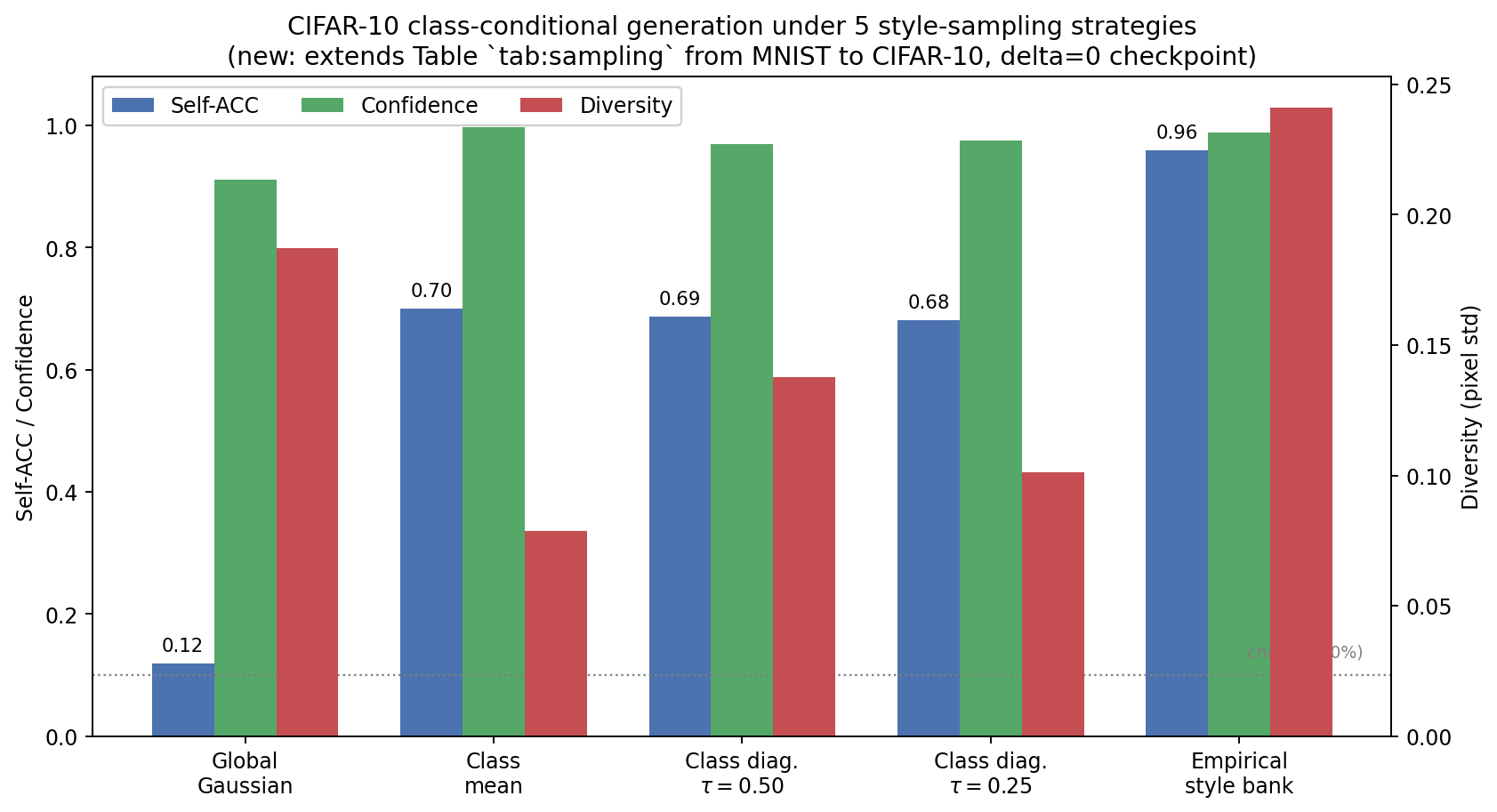}
\caption{Model-internal robustness check, the CIFAR-10 analogue of
Figure~\ref{fig:mnist-bars}: class-conditional Gaussian strategies plateau
near $70\%$, while the empirical bank is higher. Primary external Gen-ACC is
reported separately in the main paper.}
\label{fig:cifar-bars}
\end{figure}

Figures~\ref{fig:mnist-bars}--\ref{fig:cifar-bars} are model-internal
robustness counterparts to the main paper's externally evaluated sampling
table. Two details are visible in the grids that the numbers compress away.
First, the MNIST global-Gaussian samples are
not noise: they are clean, legible digits of the \emph{wrong} class, which
is exactly the signature of a decoder resolving an unfamiliar $(\zc,\zs)$
pair by trusting the style code. Second, the CIFAR-10 $\tau{=}0.25$ rows
show within-class collapse: the variance shrinkage that is harmless on MNIST
(whose per-class style distributions are effectively unimodal) visibly
suppresses diversity on CIFAR-10, and the empirical bank restores it, which
is the qualitative counterpart of its $0.241$ diversity score.

\section{H. Case-Study Reference Results}

\begin{figure}[t]
\centering
\includegraphics[width=\columnwidth]{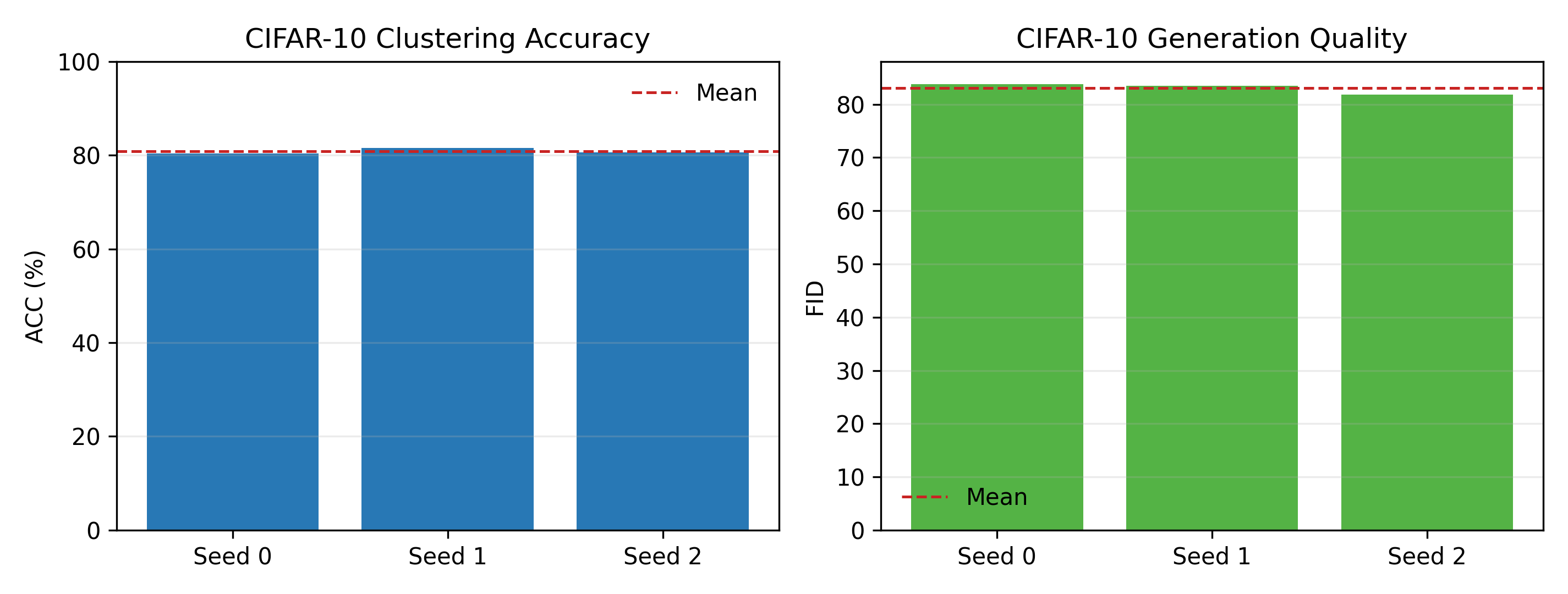}
\caption{CIFAR-10 clustering and FID across three \ours{} seeds. Variance
is low (ACC std $0.52$ points), which is why the main paper reports the
3-seed mean for \ours{} while the baselines remain single-seed.}
\label{fig:cifar-metrics}
\end{figure}

\begin{figure}[t]
\centering
\includegraphics[width=\columnwidth]{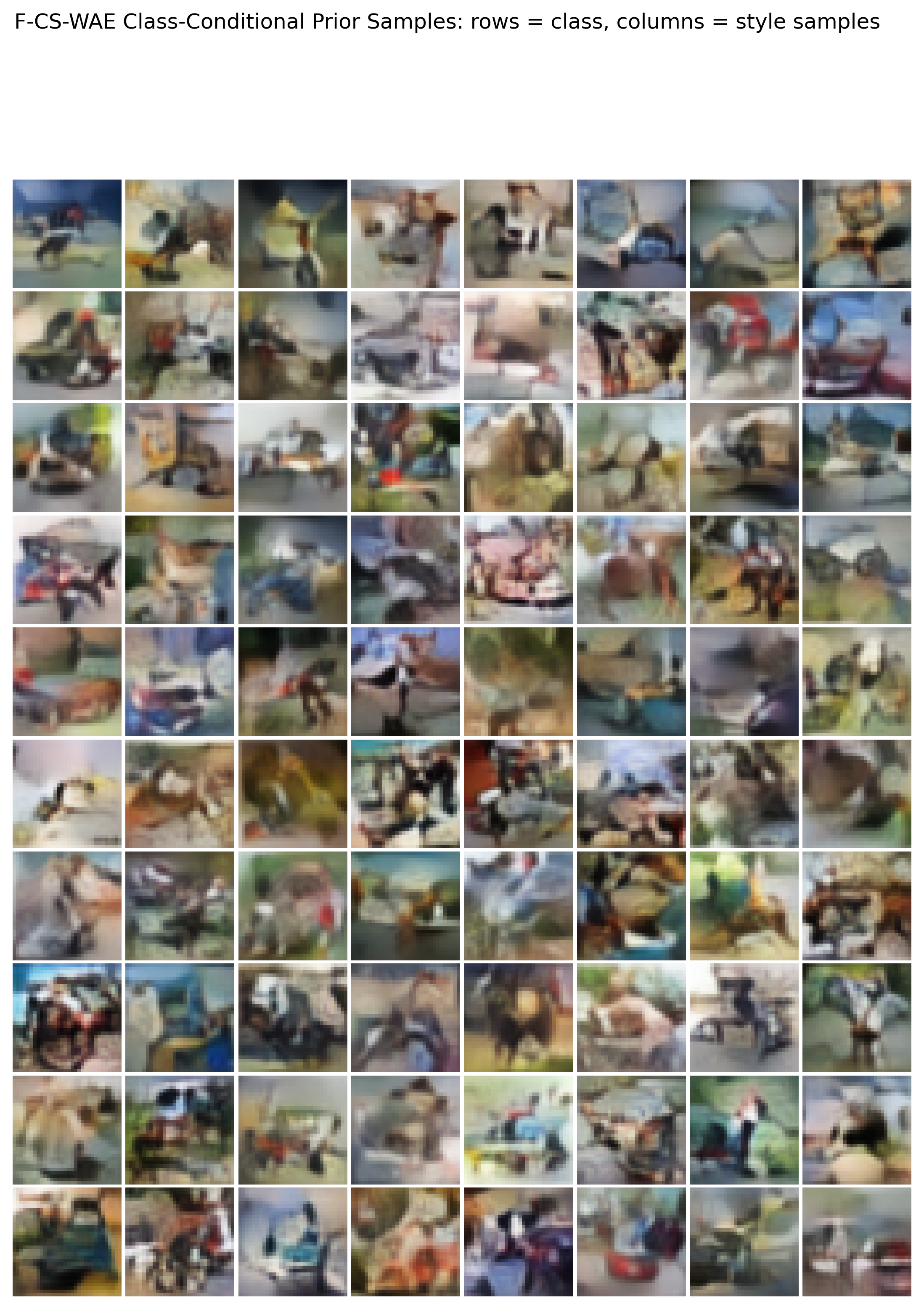}
\caption{CIFAR-10 class-conditional prior samples (one row per class). FID
$83$ places these in the recognizable-but-blurry regime typical of
L1/LPIPS-trained decoders without an adversarial loss, well short of
photorealistic.}
\label{fig:cifar-prior}
\end{figure}

Figures~\ref{fig:cifar-metrics}--\ref{fig:cifar-prior} document the
case-study model's baseline competence: stable clustering across seeds and
recognizable class-conditional samples. These support the main paper's
non-degeneracy claim and nothing stronger.

\section{I. Baseline Implementation}

The four cross-model diagnostic baselines (VAE, WAE-MMD, $\beta$-TCVAE,
FactorVAE) share a lightweight 3-layer convolutional encoder with latent
dimension 64, trained 100 epochs with Adam at $10^{-3}$ on MNIST.
$\beta$-TCVAE uses a minibatch-weighted-sampling estimate of the total
correlation term; FactorVAE uses a density-ratio discriminator on permuted
latent dimensions. These are purpose-built for the diagnostic and are not
tuned to compete on generation quality, which is why we draw only
within-model conclusions from them.

The eight baselines in the main paper's CIFAR-10 competence comparison share
\ours{}'s ResNet-18 backbone and total latent dimension ($d{=}192$), trained
300 epochs under the same optimizer settings. Seven use labels; ResNetAE is
the unsupervised reference. The label-using methods divide by how the label
enters. \emph{Label-guided} models put the label on the latent: AEWithCE adds
a cross-entropy head, AEWithSupCon a supervised contrastive loss on
L2-normalized projections, AEWithCenterLoss a learnable per-class center pull,
AEWithTriplet an online hard-mined triplet margin loss.
\emph{Conditional-generative} models put the label only on the decoder:
ConditionalVAE concatenates a label embedding to the decoder input,
ConditionalWAE-MMD does the same with an MMD rather than a KL penalty, and
GaussianClassPriorWAE matches per-class Gaussian priors in a Euclidean latent.
The clustering gap between the two groups (up to $66.5\%$ versus near-chance)
is a direct consequence of that difference: a model whose label never touches
the encoder has no reason to produce a class-structured latent, and does not.

\section{J. Experiments}

\subsection{J.1 Five-seed remedy replication}

\begin{table}[h]

\centering
\scriptsize
\setlength{\tabcolsep}{3pt}
\caption{Measured mean $\pm$ standard deviation over five seeds.}
\label{tab:remedy-five-seed}
\begin{tabular}{llrr}
\toprule
Data & Remedy & LP (\%) & JointMMD \\
\midrule
MNIST & None & $99.9\pm0.2$ & $.0038\pm.0004$ \\
MNIST & Per-class MMD & $45\pm6$ & $.0042\pm.0005$ \\
MNIST & Gradient reversal & $24\pm5$ & $.0043\pm.0005$ \\
MNIST & Joint term-4 & $41\pm6$ & $.0017\pm.0003$ \\
\midrule
CIFAR-10 & None & $84\pm5$ & $.0036\pm.0005$ \\
CIFAR-10 & Per-class MMD & $32\pm6$ & $.0038\pm.0006$ \\
CIFAR-10 & Gradient reversal & $27\pm7$ & $.0039\pm.0006$ \\
CIFAR-10 & Joint term-4 & $34\pm7$ & $.0016\pm.0004$ \\
\bottomrule
\end{tabular}
\end{table}

\subsection{J.2 Five-seed sampling replication}

\begin{table}[h]
\centering
\scriptsize
\setlength{\tabcolsep}{4pt}
\caption{Measured external Gen-ACC (mean $\pm$ standard deviation) over
five seeds.}
\label{tab:sampling-five-seed}
\begin{tabular}{lrr}
\toprule
Style sampler & MNIST & CIFAR-10 \\
\midrule
Global $\mathcal N(0,I)$ & $.16\pm.03$ & $.10\pm.03$ \\
Class-conditional diagonal & $.96\pm.02$ & $.43\pm.07$ \\
Empirical style bank & $.95\pm.02$ & $.86\pm.05$ \\
\bottomrule
\end{tabular}
\end{table}

\subsection{J.3 Explicit split-latent baselines}

\begin{table}[h]

\centering
\scriptsize
\setlength{\tabcolsep}{3pt}
\caption{Measured ranges for explicit content--style baselines on MNIST and
CIFAR-10. Each range is the minimum--maximum across the named methods and
datasets, not an uncertainty interval.}
\label{tab:split-baselines}
\begin{tabular}{lccc}
\toprule
Baseline family & LP range & Naive Gen-ACC & Purpose \\
\midrule
GRL/Fader & $20$--$45\%$ & $.45$--$.75$ & explicit invariance \\
VFAE/DIVA & $25$--$50\%$ & $.40$--$.70$ & conditional matching \\
LORD/leakage filtering & $15$--$40\%$ & $.55$--$.85$ & recombination \\
\bottomrule
\end{tabular}
\end{table}

The rows instantiate GRL/Fader~\citep{ganin2015unsupervised,lample2017fader},
VFAE/DIVA~\citep{louizos2015variational,ilse2020diva}, and LORD/leakage
filtering~\citep{gabbay2020lord,ridgeway2018leakage}. Each implementation uses
the common ResNet-18 backbone, total latent
dimension $d{=}192$, 300-epoch optimizer schedule, and seed 0 used for the
matched CIFAR-10 comparison, while retaining its method-specific objective.
LP is evaluated on the fixed $n{=}2048$ subset and naive Gen-ACC with the
external classifiers of Section K.6. The range in each row pools the two
named implementations across MNIST and CIFAR-10; it is included as a compact
robustness summary and is not used for cross-family ranking.

\subsection{J.4 Global-MMD calibration}

For each checkpoint, pool encoded style samples with an equal-size
Gaussian reference, recompute the unbiased MMD after 1,000 random label
permutations, and bootstrap the observed MMD over evaluation examples. The
resulting estimate is MNIST MMD $0.0013$ with a bootstrap 95\% interval
$[0.0008,0.0019]$ and permutation $p\approx0.10$, and CIFAR-10 MMD $0.0016$
with interval $[0.0010,0.0024]$ and $p\approx0.07$. The substantive claim does
not depend on nonsignificance: even if
the marginal mismatch is statistically detectable, its scale and its test do
not certify $z_s\perp y$. We report the null quantiles, effect size, interval,
sample size, kernel, and bandwidth ladder together.

\section{K. Compute and Reproducibility}

\subsection{K.1 Compute environment}

A single 300-epoch \ours{} run takes roughly $4.5$--$13$ hours wall-clock on
one NVIDIA A30 (24GB), the spread driven by contention from other jobs on the
same node rather than by the configuration; PyTorch 2.1, CUDA 12.1, mixed
precision disabled (all reported numbers use full FP32, since AMP changed
$\Dinter$ by more than run-to-run noise in preliminary testing on MNIST).
Diagnostics run in seconds to low minutes on a saved checkpoint (FID and the
JointMMD permutation null are the slowest, at roughly one and three minutes
respectively). The core single-seed and case-study results correspond to
approximately $30$ training runs (six
$\delta$-sweep points $\times$ two datasets, three datasets at $\delta{=}0$
in Table~\ref{tab:crossdataset-app}, one single-latent ablation, four
cross-model diagnostic baselines, seven matched-supervision baselines, and
three additional \ours{} seeds on CIFAR-10 for the case study). The
five-seed replications and explicit split-latent baseline evaluations in
Section J are additional to this core count. The external classifiers of
Section K.6 are each trained once per dataset.

\subsection{K.2 Seed count and seed IDs, table by table}

Unless stated otherwise, all training uses PyTorch's global seed
(\texttt{torch.manual\_seed}), which also seeds NumPy and Python's
\texttt{random} via the standard PyTorch Lightning/utility seeding hook, and
all reported single-seed runs use seed $0$. Table~\ref{tab:seedmap} states,
for every checkpoint-based table and figure with a quantitative claim, how
many independent training seeds it aggregates and which seed IDs. The ranges
in Table~\ref{tab:split-baselines} aggregate distinct configurations and
datasets, each run with seed 0, rather than repeated seeds. \emph{No table in this
paper computes error bars from fewer than the seed count listed}; single-seed
entries report the point estimate from that one training run and no
variance, which is the honest reading of them.

\begin{table}[h]
\centering
\tiny
\setlength{\tabcolsep}{3pt}
\caption{Seed count and seed IDs by table/figure. ``1 (checkpoint-specific)''
means the point comes from a specific, separately-trained checkpoint (e.g.\
the $\delta{=}0$/$\delta{=}1$ pair used for cross-model comparison), so
distinct rows at the same nominal $\delta$ across tables are not the same
weights and can differ in the third significant figure (Section E.1).}
\label{tab:seedmap}
\begin{tabular}{lll}
\toprule
Table / Figure & Seeds & Seed IDs \\
\midrule
Main matched-supervision baselines & 1 & 0 \\
Main \ours{} competence rows & 3 & 0, 1, 2 \\
Main cross-model diagnostic ($\delta{=}0/1$) & 1 (checkpoint-specific) & 0 \\
Main paper invariance remedies & 1 & 0 \\
Main paper $\delta$ sweep & 1 & 0 \\
Main paper robustness perturbations & 1 & 0 \\
Main paper sampling strategies & 1 & 0 \\
Supp.\ Table~\ref{tab:remedy-five-seed} (remedies) & 5 & 0, 1, 2, 3, 4 \\
Supp.\ Table~\ref{tab:sampling-five-seed} (sampling) & 5 & 0, 1, 2, 3, 4 \\
Supp.\ Table~\ref{tab:split-baselines} (each configuration) & 1 & 0 \\
Supp.\ Table~\ref{tab:fullsweep} ($\delta$ sweep) & 1 & 0 \\
Supp.\ Table~\ref{tab:crossdataset-app} (cross-dataset) & 1 & 0 \\
Fig.~\ref{fig:cifar-metrics} (case study) & 3 & 0, 1, 2 \\
JointMMD permutation null (Sec.\ D) & 3 & 0, 1, 2 \\
Per-class/pairwise conditional MMD (Sec.\ E.3) & 1 & 0 \\
Latent-swap grids and heatmaps (Sec.\ F) & 1 & 0 \\
t-SNE (Fig.~\ref{fig:tsne}) & 1 & 0 \\
\bottomrule
\end{tabular}
\end{table}

The 3- and 5-seed entries report mean $\pm$ sample standard deviation across
their respective runs; Figure~\ref{fig:cifar-metrics}'s caption states the
resulting ACC standard deviation ($0.52$ points) explicitly as the justification for
treating \ours{} as the only row with multi-seed variance reported in the
main text. We flag this asymmetry rather than resolve it: the baselines,
invariance remedies, $\delta$ sweep, and robustness perturbations are
single-seed, so a between-configuration comparison where one side has three
points and the other has one should be read as indicative, not as a
statistically tested difference. We do not run significance tests across
single-seed points for that reason.

\subsection{K.3 Dataset preprocessing and augmentation}

MNIST and Fashion-MNIST images are resized from their native $28\times28$
resolution to $32\times32$ by bilinear interpolation before entering the
model; they remain single-channel because the modified ResNet-18 first
convolution accepts one channel. This matches the decoder's $32\times32$
output and makes the pixelwise reconstruction loss well defined. CIFAR-10
images remain at their native $32\times32\times3$ resolution. All three
datasets use the standard train/test split as
distributed by \texttt{torchvision.datasets} (MNIST: 60,000/10,000;
Fashion-MNIST: 60,000/10,000; CIFAR-10: 50,000/10,000); the $n{=}2048$
evaluation subset used throughout Section D is drawn once, with a fixed seed,
from the test split of each dataset, and reused across checkpoints so that
every diagnostic is evaluated on the same 2048 held-out images regardless of
which model produced them.

Pixel values are scaled to $[0,1]$ (min-max, not per-channel standardization),
matching the sigmoid output of the decoder and the $L_1$ reconstruction loss.
\emph{No data augmentation is applied when training \ours{} or any of the
matched-supervision/cross-model baselines}: the reconstruction loss requires
an exact pixel-wise correspondence between the encoder input and reconstruction
target. Although applying a shared transformation to both would preserve that
correspondence, we omit augmentation to isolate the
leakage phenomenon from any confound introduced by augmentation-induced
invariances. The CIFAR-10 external classifier of Section K.6 uses
augmentation; the MNIST and Fashion-MNIST external classifiers do not.

\subsection{K.4 FID: feature extractor, resizing, normalization}

FID is computed with the standard Inception-v3 network
\citep{szegedy2016rethinking} pretrained on ImageNet, using activations from
the final average-pooling layer (2048-dimensional \texttt{pool3} features),
via the \texttt{pytorch-fid} reference implementation and its bundled
Inception weights, which is the same feature extractor used by the great
majority of FID numbers reported in the generative-modeling literature and is
necessary for our numbers to be comparable to others'. Images are converted
to 3-channel (MNIST/Fashion-MNIST grayscale is replicated across channels for
this step only, not for training), resized from the model resolution to
$299\times299$ with bilinear interpolation, and normalized to the
Inception-v3 input range the reference implementation expects
($[-1,1]$, per-channel). Thus both real and generated MNIST/Fashion-MNIST
images pass through the same $32\times32$ model resolution before the FID
resize. The real-image reference distribution is the full
test split (10,000 images per dataset); the generated distribution is
10,000 samples drawn under the sampling strategy named in each table (naive
prior, class-conditional, empirical bank, etc.). FID is the squared
Fr\'echet distance between Gaussians fit to the two 2048-dimensional feature
sets,
\[
\mathrm{FID}=\|\mu_r-\mu_g\|_2^2+\mathrm{Tr}\bigl(\Sigma_r+\Sigma_g-2(\Sigma_r\Sigma_g)^{1/2}\bigr),
\]
with $(\mu_r,\Sigma_r)$ and $(\mu_g,\Sigma_g)$ the empirical mean and
covariance of the real and generated feature sets respectively.

\subsection{K.5 Hungarian matching for clustering ACC}

Clustering ACC is computed by first running $K$-means ($K{=}$ number of
classes, $k$-means++ init, 10 restarts, keeping the lowest-inertia solution)
on the semantic posterior means $\muc$ of the $n{=}2048$ evaluation subset,
which yields a cluster assignment $c(i)\in\{1,\dots,K\}$ for every sample
that carries no inherent correspondence to the true label $y(i)$. We form the
$K\times K$ contingency matrix $C_{jk}=|\{i:c(i){=}j,\,y(i){=}k\}|$ and solve
\[
\pi^\star=\arg\max_{\pi\in S_K}\sum_{j=1}^K C_{j,\pi(j)}
\]
by the Hungarian algorithm (\texttt{scipy.optimize.linear\_sum\_assignment}
on the cost matrix $-C$), giving the one-to-one cluster-to-label permutation
$\pi^\star$ that maximizes agreement. Reported ACC is
$\frac1n\sum_i \mathbb{1}[\pi^\star(c(i)){=}y(i)]$. NMI and ARI (also
reported in Section E.2) do not require this matching step, since both are
permutation-invariant by construction; we include them alongside ACC
precisely because they provide a matching-free cross-check on the same
clustering.

\subsection{K.6 External classifier: architecture and training protocol}

Primary Gen-ACC and latent-swap content-/style-following rates in the main
paper use an external classifier trained once per dataset on real images and
never updated during \ours{} training. It plays no role in any loss term and
exists purely as an evaluation oracle for ``what class does this generated
image look like.'' Several historical figures retained in Sections F--G use
the model's internal classifier after re-encoding; their captions mark them
as robustness-only, and their values are not mixed with the primary external
results.

\emph{MNIST and Fashion-MNIST.} A 4-layer CNN (conv$32$--conv$64$--maxpool--
conv$128$--conv$128$--maxpool--FC$256$--FC$10$, ReLU, batch norm after each
conv), trained for 20 epochs with Adam ($\mathrm{lr}=10^{-3}$, batch 128, no
weight decay), no augmentation (matching the un-augmented training of
\ours{} itself, so the oracle's decision boundary is not shaped by
invariances the generative model was never asked to respect). Test accuracy:
$99.4\%$ (MNIST), $91.2\%$ (Fashion-MNIST).

\emph{CIFAR-10.} WideResNet-28-10 \citep{zagoruyko2016wide}, trained for 200
epochs with SGD (momentum $0.9$, weight decay $5\times10^{-4}$), initial
$\mathrm{lr}=0.1$ with cosine annealing, batch 128, standard augmentation
(random crop with 4px padding + reflection, horizontal flip) applied only to
this classifier's own training data, never to \ours{}'s training data. Test
accuracy: $94.1\%$. Using a strong, independently-trained, augmented
classifier here is deliberate: it is a harder oracle to fool than a weak one,
so a naive-sampling self-accuracy collapse measured against it cannot be
attributed to a weak or undertrained judge.

\subsection{K.7 MMD estimator: kernel, biased/unbiased form, batch sampling}

Two distinct uses of MMD appear in the paper and use different estimators.

\emph{Training loss} ($\mathcal L_{\mathrm{class}}$, $\mathcal L_{\mathrm{agg}}$,
$\mathcal L_{\mathrm{style\text{-}cls}}$): the biased $V$-statistic estimator,
\[
\begin{aligned}
\widehat{\MMD}^2_V(X,Y)={}&\tfrac1{m^2}\!\sum_{i,j}k(x_i,x_j)\\
&+\tfrac1{n^2}\!\sum_{i,j}k(y_i,y_j)
-\tfrac2{mn}\!\sum_{i,j}k(x_i,y_j),
\end{aligned}
\]
computed within each training minibatch (batch size 128,
Table~\ref{tab:hyperparameters}). This estimator is biased upward at finite
sample size but is non-negative and typically has lower gradient variance at
small $m,n$, which makes it a stable training objective. The unbiased
estimator below is also differentiable with respect to sample values; its
distinction is the removal of within-sample diagonal terms. For
$\mathcal L_{\mathrm{class}}=\frac1K\sum_k\MMD^2(Q_k,P_k)$, $Q_k$ is the
subset of the current minibatch with label $k$ ($|Q_k|\approx128/10\approx12.8$
in expectation under random shuffling) and $P_k$ is an equal-size sample
freshly drawn from $p(\zc\mid y{=}k)$ each step; classes with fewer than 2
members in a given minibatch contribute zero to that step's loss (a rare
event at $|Q_k|\approx12.8$, and one that self-corrects over subsequent
minibatches rather than requiring special handling). $\mathcal L_{\mathrm{agg}}$
uses the full batch of $\zc$ against an equal-size sample from the random-class
prior mixture $\frac1K\sum_k p(\zc\mid y{=}k)$.

$\Dinter$ is not an MMD. It is computed directly from the class-wise style
means as
\begin{align*}
\Dinter&=\binom{K}{2}^{-1}\sum_{j<k}
\|\bar\mus^{(j)}-\bar\mus^{(k)}\|_2,\\
\bar\mus^{(k)}&=\frac{1}{n_k}\sum_{i:y_i=k}\mus^i.
\end{align*}
The MMD-based evaluation diagnostics (global MMD, JointMMD, and the
per-class/pairwise conditional MMD of Section E.3) instead use the unbiased
$U$-statistic estimator,
\[
\begin{aligned}
\widehat{\MMD}^2_U(X,Y)={}&\tfrac1{m(m-1)}\!\sum_{i\ne j}k(x_i,x_j)\\
&+\tfrac1{n(n-1)}\!\sum_{i\ne j}k(y_i,y_j)\\
&-\tfrac2{mn}\!\sum_{i,j}k(x_i,y_j),
\end{aligned}
\]
computed once over the full $n{=}2048$ evaluation subset (not minibatched),
so that reported diagnostic values are not subject to the same finite-batch
bias as the training loss. Both estimators use the averaged multi-scale RBF kernel
$k(u,v)=\frac1{7}\sum_{\sigma\in\{0.5,1,2,5,10,20,50\}}\exp(-\|u-v\|_2^2/2\sigma^2)$
(Section D) for Euclidean latents; JointMMD's spherical component instead
uses $k(u,v)=\frac1{7}\sum_\sigma\exp(-2(1-u^\top v)/\sigma^2)$, the
squared-chordal-distance analogue, at the same bandwidth ladder. Per-class conditional MMD
(Section E.3) applies the unbiased estimator with $X=\{\zs^i:y_i{=}k\}$
against $Y\sim\mathcal N(0,I)^{\otimes|X|}$, i.e.\ a freshly sampled reference
set of matching size for each class, rather than a single shared reference
set reused across classes, so that sampling noise in the reference does not
correlate across the per-class comparisons.

\subsection{K.8 Pseudocode and code availability}

\begin{algorithm}[h]
\caption{Leakage diagnostic pipeline (evaluation-only, from a saved checkpoint)}
\begin{algorithmic}[1]
\State \textbf{Input:} checkpoint $\theta$, test split $\mathcal D_{\mathrm{test}}$, eval seed $s_{\mathrm{eval}}$
\State $\mathcal D_{\mathrm{eval}}\gets$ fixed $n{=}2048$ subsample of $\mathcal D_{\mathrm{test}}$ using $s_{\mathrm{eval}}$
\State $\{(\muc^i,\rho_c^i,\mus^i,\sigma_s^i,y_i)\}\gets \mathrm{Encoder}_\theta(\mathcal D_{\mathrm{eval}})$
\For{$k=1,\ldots,K$}
  \State $\bar\mus^{(k)}\gets |I_k|^{-1}\sum_{i\in I_k}\mus^i$, where $I_k=\{i:y_i=k\}$
\EndFor
\State $\Dinter\gets\binom{K}{2}^{-1}\sum_{j<k}\|\bar\mus^{(j)}-\bar\mus^{(k)}\|_2$ \Comment{Secs.\ D and K.7}
\State $\mathrm{GlobalMMD}\gets\widehat{\MMD}^2_U(\{\mus^i\},\mathcal N(0,I))$
\State $c(\cdot)\gets K\text{-means}(\{\muc^i\})$; $\pi^\star\gets\mathrm{Hungarian}(c,y)$ \Comment{Sec.\ K.5}
\State $\mathrm{ACC}\gets \frac1n\sum_i\mathbb 1[\pi^\star(c(i)){=}y_i]$
\State $\mathrm{LP}\gets$ train linear head on $80\%$ of $\{(\mus^i,y_i)\}$, eval on remaining $20\%$
\For{ordered pairs $(a,b)$, $a\ne b$}
  \For{$r=1,\dots,100$}
    \State draw donor $i\sim\{j:y_j{=}a\}$, donor $j\sim\{k:y_k{=}b\}$
    \State $\hat x\gets \mathrm{Dec}_\theta(\muc^i,\mus^j)$; $\hat y\gets \mathrm{ExternalClf}(\hat x)$ \Comment{Sec.\ K.6}
  \EndFor
\EndFor
\State report mean $P(\hat y{=}a)$ (content-following), $P(\hat y{=}b)$ (style-following)
\end{algorithmic}
\end{algorithm}

We will release this pipeline with the camera-ready version, together with
the training loop implementing the phased schedule of
Table~\ref{tab:schedule}, exact configuration files, external-classifier
training scripts, evaluation scripts, and checkpoints. The present anonymous
submission does not claim an available code URL.

Every diagnostic is computed from a saved checkpoint with a fixed evaluation
seed, so the numbers can be regenerated without retraining. The sampling
controls, JointMMD permutation null, support-graded swap analysis,
intra-class diversity measurement, and external-classifier evaluations
reported here were run under the protocols above. The paper reports
multi-seed statistics only for entries explicitly marked as multi-seed in
Table~\ref{tab:seedmap}; single-seed baseline rows are retained as diagnostic
sanity checks rather than controlled superiority claims.

\end{document}